\documentclass{ar-1col-S2O}

\usepackage{amsmath,amsfonts}
\usepackage{algorithmic}
\usepackage{algorithm}
\usepackage{array}
\usepackage{lmodern}
\usepackage{textcomp}
\usepackage{stfloats}
\usepackage{verbatim}
\usepackage{graphicx}
\usepackage{nomencl}
\usepackage{ntheorem}   
\usepackage{lipsum}     
\usepackage{graphicx}
\usepackage{bm}

\DeclareRobustCommand{\uvec}[1]{{%
		\ifcsname uvec#1\endcsname
		\csname uvec#1\endcsname
		\else
		\bm{\mathbf{#1}}%
		\fi
}}

\usepackage{hyperref}

\usepackage[numbers]{natbib}
\usepackage{url}
\usepackage[titletoc]{appendix}
\jname{Xxxx. Xxx. Xxx. Xxx.}
\jvol{AA}
\jyear{YYYY}
\doi{10.1146/((please add article doi))}

\begin{document}

\markboth{Tafrishi et al.}{Path Planning Problem of Rolling Contacts}

\title{A Survey on Path Planning Problem of Rolling Contacts: Approaches, Applications and Future Challenges}

\author{Seyed Amir Tafrishi,$^1$ Mikhail Svinin,$^2$ and Kenji Tahara$^3$
\affil{$^1$Geometric Mechanics and Mechatronics in Robotics (gm$^2$R) Laboratory, School of Engineering, Cardiff University, Queen's Buildings, Cardiff, United Kingdom; email: Tafrishisa@cardiff.ac.uk}
\affil{$^2$Intelligent Robotic Systems Laboratory, Information Science
and Engineering Department, Ritsumeikan University, Shiga, Japan; email: Svinin@fc.ritsumei.ac.jp}
\affil{$^3$Human-Centered Robotics (HCR) Laboratory, Mechanical Engineering Department, Kyushu University, Fukuoka, Japan; e-mail: Tahara@mech.kyushu-u.ac.jp}}

\begin{abstract}
This paper explores an eclectic range of path-planning methodologies engineered for rolling surfaces. Our focus is on the kinematic intricacies of rolling contact systems, which are investigated through a motion planning lens. Beyond summarizing the approaches to single-contact rotational surfaces, we explore the challenging domain of spin-rolling multi-contact systems. Our work proposes solutions for the higher-dimensional problem of multiple rotating objects in contact. Venturing beyond kinematics, these methodologies find application across a spectrum of domains, including rolling robots, reconfigurable swarm robotics, micro/nano manipulation, and nonprehensile manipulations. Through meticulously examining established planning strategies, we unveil their practical implementations in various real-world scenarios, from intricate dexterous manipulation tasks to the nimble manoeuvring of rolling robots and even shape planning of multi-contact swarms of particles. This study introduces the persistent challenges and unexplored frontiers of robotics, intricately linked to both path planning and mechanism design. As we illuminate existing solutions, we also set the stage for future breakthroughs in this dynamic and rapidly evolving field by highlighting the critical importance of addressing rolling contact problems.
\end{abstract}

\begin{keywords}
path planning, contact kinematics, spin-rolling motions, rolling robotics, geometric mechanics 
\end{keywords}
\maketitle

\tableofcontents

\section{Introduction}
\label{Sec:Introduction}
Rolling is a fundamental motion that plays a pivotal role in generating locomotion \cite{jurdjevic1993geometry,laumond1998robot,montana1988kinematics}. Consider an object or particle's velocity $\uvec{v}$, which can be described by
\begin{equation}
\uvec{v} = \uvec{v}'+ \bm{\omega} \times \mathbf{r},
\label{Eq:basicKinematics}
\end{equation}
where the term $(\bm{\omega} \times \mathbf{r})$ elegantly captures the conversion of angular orientation into linear velocity as well as $\uvec{v}'$ is the drift term as a linear velocity, which can be due to physical slippage or other properties. This formulation is not confined to a specific domain; it underpins the kinematics of systems as diverse as rolling robots \cite{jurdjevic1993geometry}, car systems \cite{laumond1998robot,laumond1994motion} and even object manipulation \cite{kiss2002motion}. The universality of Equation (\ref{Eq:basicKinematics}) renders it a foundational concept that gives rise to numerous challenges and opportunities in the domains of motion planning, control, and robotics. This can be used to present the kinematics of different locations (inertial frame) of an object, either in the centre of the body or at the contact point with the ground, which is important in rolling contact kinematics.

The motion planning problem of rolling contact point that is common between at least two objects \cite{montana1988kinematics} presents two significant challenges that have deeply engaged both the robotics and control theory communities:

\textit{i. Non-holonomic Constraints:} The non-holonomic constraints arise from conditions like no-slipping or limitations in rotational angles, resulting in equations that resist straightforward integration \cite{jurdjevic1993geometry,laumond1994motion,li1990motion,jurdjevic1995non}. This introduces substantial complexities in control strategies, necessitating specialized motion planning approaches to realize feasible and optimal trajectories \cite{jurdjevic1995non}. The problem becomes more complicated when dynamic factors, depending on the robotic application under consideration, must be solved together with rolling contact kinematics \cite{Oriolo2005Feedback,morinaga2014motion}. This introduces coupling mathematical terms which impose indirect constraints on the developed approaches.
    \begin{figure}[t!]
	\centering
	\includegraphics[scale=0.7]{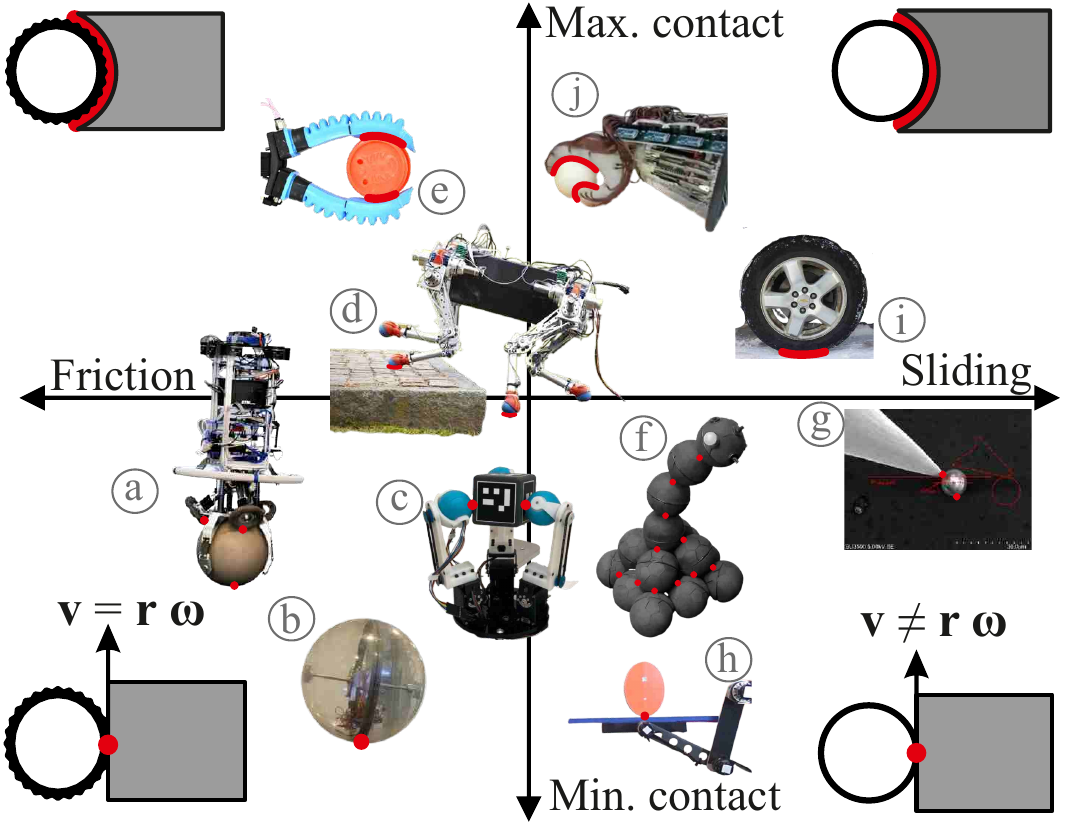}
	\caption{Different characteristics in rolling contact problem. Applications: a) ball-bot robot \cite{fankhauser2010modeling}, b) rolling spherical robots \cite{armour2006rolling}, c) agile grasping \cite{yuan2020designIROS}, d) foot placement in legged robots \cite{hutter2014quadrupedal}, e) power grasping \cite{teeple2020multi}, f) reconfigurable rolling robots \cite{{zhong2022kin}}, g) mico-/nano-particle manipulation \cite{sumer2008rolling}, h) nonprehensile manipulation \cite{woordrufLynch2023TRO}, i) wheels in mobile robot/cars \cite{tafrishi2022novel}, j) Soft hand grasping and manipulation \cite{gilday2023predictive}. }
\label{Fig:TheFingertipManipulationPathPlanningStatement}
\end{figure}

\textit{ii. Topological Complexity:} Depending on the physical and mechanical characteristics, such as the degree of freedom (DoF), the degree of rotation in the moving bodies varies. Here, we define the degree of rotation as a sub-category of DoF, focusing solely on the number of angular orientations of in-contact objects. Studies in motion planning are dedicated to addressing the intricacies associated with planning trajectories for rolling objects with different degrees of rotation \cite{jurdjevic1993geometry,li1990motion,Oriolo2005Feedback,date2004simultaneous,svinin2008motion,tafrishi2023geometric}. However, the topological properties of multi-contact rolling surfaces make analytical solutions dramatically challenging \cite{kiss2002motion,sarkar1996velocity}. This challenge arises from computational complexity, including arbitrary convex surfaces that require the integration of differential geometric formulations \cite{montana1988kinematics,woodruff2019second}.

While these challenges are inherently interconnected, the approaches developed to address each issue can, at times, lead to their distinct treatment. Effectively managing non-holonomic constraints demands specialized control techniques, whereas motion planning strategies focus on devising algorithms and methodologies to determine optimal paths and trajectories for rolling objects with a specific degree of rotation. Our goal is to distill the complex mathematical aspects of motion planning strategies at the kinematic level, making them practically relevant for robotics applications. In robotics, path planning and trajectory tracking are fundamentally distinct concepts \cite{laumond1998robot}, particularly in rolling contact systems. Trajectory tracking (including point-to-point convergence) focuses on stabilizing a system's dynamics to follow a predefined and realizable reference, whereas path planning generates feasible paths while addressing constraints and multi-solution complexities unique to rolling dynamics. In rolling systems, not all paths are dynamically achievable due to nonholonomic constraints, the coupling of rotation and translation, and the contact kinematics shaped by surface changes. Poorly planned paths can lead to instability or unachievable motion, affecting overall dynamics and complicating controller stability and robustness \cite{morinaga2014motion,ozawa2017grasp}. This survey bridges the gap by providing insights into effective path planning tailored to rolling contact systems, while briefly addressing practical challenges in achieving stable motion—widely studied in control theory. By comprehensively understanding open problems and existing challenges, the field of robotics and control (examples are depicted in Figure \ref{Fig:TheFingertipManipulationPathPlanningStatement}) could benefit from more efficient and versatile solutions for robotic mechanisms and systems involving path planning of rolling contact.

The applications of studies on rolling have evolved significantly since its inception, originally rooted in investigations of car-like systems (rolling disks), primarily focused on achieving specific vehicle configurations \cite{laumond1998robot}. As researchers \cite{li1990motion} delved deeper into Montana kinematics \cite{Montana1988} with generalised rolling contact between two objects, the scope expanded to address increasingly complex problems in robotics. The emergence of rolling robots \cite{armour2006rolling,robotics1010003,Minggang2023} and ball-bots  \cite{lauwers2006dynamically} represented a milestone, capturing the interest of roboticists in leveraging this property for motion. The exploration of additional DoF subsequently shifted focus towards dexterous manipulation and grasping tasks \cite{tahara2012externally}.  Despite initially seeming clear, these applications converge to the fundamental challenge of the rolling contact problem within the kinematic domain. Presently, more intricate issues have surfaced, incorporating concepts from differential geometry and topology. This facilitates the creation of multiple dynamic 2D manifolds with potentially evolving curvature over time, enabling the integration of deformable objects for tasks ranging from in-hand manipulation \cite{tahara2012externally,sanchez2018robotic} to coordinating swarm robotic systems \cite{zhong2022kin}. This evolution underscores the continual refinement and expansion of the rolling contact problem's domain of application within the field of robotics, emphasizing the importance of geometric mechanics in understanding the root of robotics problems rather than considering systems as black-box entities.

In this paper, we highlight the challenges of rolling contact within the robotics community, emphasizing its interdisciplinary significance across engineering and mathematics. One such challenge is the intricacy of motion planning for non-uniform object formats, where real-time changes in surface curvature demand a thorough understanding of trajectory dynamics. The inherently underactuated nature of rolling contact kinematics, governed by no more than three inputs and five states (two per object in contact and one indicating the relative angle between them), imposes significant constraints on planned motion, dictating how motion unfolds. Integrating rolling contacts with the broader dynamics of the system or robot presents an additional complex challenge, underscoring the multifaceted nature of this research problem. Furthermore, we address how the fundamental problem persists, even in more complex scenarios involving multi-contact surfaces.

This survey paper serves as a bridge between the literature of geometric mechanics studies in rolling contacts and the challenges encountered in applied robotics, offering theoretical solutions for unresolved issues. The contributions of this work include:
\begin{summary}[SUMMARY OF CONTRIBUTIONS]
\begin{itemize}
\item Discussion of the rolling contact kinematics problem, considering contact types and properties, and the development of generalized Montana kinematics for multi-contact spin-rolling surfaces (Section \ref{Sec:GeoemtricKinemtatics}),
\item Presentation of state-of-the-art motion planning solutions for various rolling problems, including disks, pure rolling, and spin-rolling motions, with an extension to multi-contact points and proposals for hypothetical solutions (Section \ref{Sec:PathPlanning}),
\item Spotlighting application-based studies that connect with fundamental mathematical problems, articulating remaining challenges in geometric mechanics and paving the way for diverse robotics applications (Section \ref{Sec:RoboticApplications}).
\end{itemize}
\end{summary}

\section{Geometry \& Kinematics of Rolling Contact Problem}
\label{Sec:GeoemtricKinemtatics}
In this section, we study the kinematic properties inherent to rolling contact and address the challenges of associated constraints. To facilitate our investigation, we extend the Montana kinematics framework \cite{montana1988kinematics} to encompass multi-contact surfaces, marking a pioneering step in this direction. This generalization proves instrumental in tackling various robotics problems with different mechanical or mechanism designs.
\subsection{Common Contact Points Between Rolling Surfaces}
Rolling contact primarily involves interactions between a common point between two rolling surfaces. This contact between two surfaces can be categorized into four cases based on friction and the characteristics of the contact surface, as shown in Figure \ref{Fig:TheFingertipManipulationPathPlanningStatement}. In most research scenarios, it is assumed that the rolling objects in contact have rough surfaces (reasonable Coulomb friction), conforming to the \textit{no-sliding} constraint where $\uvec{v} = \bm{\omega} \times \uvec{r}$ at the contact point. Many robotics studies from an application perspective focus on the left side of the graph depicted in Figure \ref{Fig:TheFingertipManipulationPathPlanningStatement}, due to the consideration of rough surfaces (high contact point Columnb friction). However, a significant challenge arises when slippage occurs at the contact point and the surfaces of the objects are smooth, either due to material properties or nonprehensile motion actions. This introduces an additional drift term into the contact kinematics, as shown in (\ref{Eq:basicKinematics}).  Managing drift terms from a control perspective is relatively straightforward when states are bounded. For instance, if the drift term is \textit{weakly positively Poisson stable} \cite{manikonda1997controllability,wang2011note} with satisfied \textit{Lie algebra rank condition (LARK)} for local accessibility of the complete system, we can consider the rolling contact kinematics with slippage controllable. Nonetheless, understanding how to sense, model and control slippage in practical application remains a paramount challenge in the field. For instance, Jia proposed a planning strategy for a slipping sphere with an initial velocity, following a parabolic trajectory \cite{jia2016planning}. This work also accentuated that while planning for rolling objects can often be conducted at the kinematic level, addressing slippage, which can generate agile motions, necessitates a meticulous approach and the development of control mechanisms for managing slippage velocities. When slippage is induced with random variations, such as an unknown initial velocity, planning becomes more intricate as it must be compensated for with limited rotational inputs $\bm{\omega}$, assuming that we can determine the friction constant at the contact point \cite{shi2017dynamic}.

The area of contact significantly influences the kinematic characteristics, encompassing both physical and geometrical properties. From a geometric perspective, the meticulous definition of curvature at the contact point is pivotal, a matter we will delve into in subsequent sections. On the physical front, there exists a misconception that enlarging the contact area, particularly with elastic objects, would streamline the control of rolling objects by increasing friction points. However, reality diverges from this assumption \cite{boucly2007modeling}. In scenarios where slippage occurs, the kinematics of the contact area undergo alteration, resulting in a nuanced model rather than the anticipated simplified friction cone model \cite{fakhari2019modeling,romeo2020methods}. This nuanced model introduces novel phenomena, such as partial-slip with stick-slip models \cite{paggi2014partial}. These complexities underscore the intricate interplay between contact area dynamics and the overall kinematic behaviour of rolling objects.

\subsection{Generilized Rolling Contact Kinematics}
The kinematics of rolling contact are contingent on the curvature and rotational states of the rolling objects. In here, we delve into the development of contact kinematics, followed by an overview of generalized kinematics and their associated properties. This foundation will pave the way for our subsequent discussion on the motion planning problem.

Numerous literature reviews in rolling contact kinematics have been based on specific definitions and frames of reference. Neimark and Fufaev initially addressed velocity equations for a frame moving along curved lines \cite{ne_mark2004dynamics}. This was followed by studies focused on deriving contact kinematics of rigid bodies in planar \cite{cai1986planar} and specialized cases \cite{cai1987spatial}. Montana introduced the widely recognized contact kinematics, which leverages the curvatures of two rotational objects with sliding terms \cite{montana1988kinematics}. This generalization was pivotal in understanding how curvature properties, including normal curvature, geodesic curvature, and torsion, influence rotational motion velocities. Li and
Canny \cite{li1990motion} using the Montana kinematics demonstrated the plate-ball system is controllable and the same condition holds for two rotating spheres with different radii. Sarkar extended Montana kinematics by incorporating acceleration terms \cite{sarkar1996velocity}. Woodruff and Lynch extended the second-order equations to accommodate complex surfaces, such as ellipsoids \cite{woodruff2019second}. However, this mathematical derivation of rolling contact kinematics comes with certain limitations. Obtaining equations can be challenging, and it is sensitive to changes in local parameters. Furthermore, its differentiability is restricted, typically up to the second order, due to explicit dependence on the Christoffel symbols and their time derivatives. 
\begin{figure*}[t!] 
\begin{eqnarray}
&\left[\begin{array}{c}
\dot{\uvec{u}}_{o,i}\\
\dot{\uvec{u}}_{f,i}\\
\psi_i
\end{array}\right]= \left[\begin{array}{ccc}
-\uvec{M}_{oi}^{-1}\left(\uvec{K}_{oi}+\tilde{\uvec{K}}_{fi}\right)^{-1} \uvec{E}_2 & \uvec{0}   & \uvec{0} \\
\uvec{0}  & -\uvec{M}_{fi}^{-1}\uvec{R}_{\psi,i} \left(\uvec{K}_{oi}+\tilde{\uvec{K}}_{fi}\right)^{-1}\uvec{E}_2 & \uvec{0} \\
-\uvec{T}_{oi}\left(\uvec{K}_{oi}+\tilde{\uvec{K}}_{fi}\right)^{-1} \uvec{E}_2 & -\uvec{T}_{fi}\uvec{R}_{\psi,i} \left(\uvec{K}_{oi}+\tilde{\uvec{K}}_{fi}\right)^{-1}\uvec{E}_2  &  -\uvec{E}_1
\end{array}\right] \bm{\omega}_{rel,i} \nonumber\\
& + \left[\begin{array}{ccc}
\uvec{M}_{oi}^{-1}\left(\uvec{K}_{oi}+\tilde{\uvec{K}}_{fi}\right)^{-1} \tilde{\uvec{K}}_f  \uvec{E}_3 & \uvec{0}   & \uvec{0}  \\
\uvec{0}  & \uvec{M}_{fi}^{-1}\uvec{R}_{\psi,i} \left(\uvec{K}_{oi}+\tilde{\uvec{K}}_{fi}\right)^{-1}  \uvec{K}_{oi}  \uvec{E}_3 & \uvec{0} \\
\uvec{T}_{oi}\left(\uvec{K}_{oi}+\tilde{\uvec{K}}_{fi}\right)^{-1} \tilde{\uvec{K}}_f \uvec{E}_3 & \uvec{T}_{fi}\uvec{R}_{\psi,i} \left(\uvec{K}_{oi}+\tilde{\uvec{K}}_{fi}\right)^{-1} \uvec{K}_{oi}\uvec{E}_3  & \uvec{0} 
\end{array}\right] \uvec{v}_{rel, i} 
\label{Eq:Montanakinemitcstransformed}
\end{eqnarray} 
\end{figure*}

Recent works have focused on parameterizing contacted two rigid no-sliding bodies' motion in different domains to facilitate easier motion control without direct integration, utilizing Darboux frame kinematics \cite{cui2010darboux,cui2017hand}. This approach formulates the revised kinematics in terms of three contravariant vectors and geometric invariants: the arc length of the contact trajectory curve and the induced curvatures of the two surfaces \cite{cui2010darboux} or sandwiched multiple \cite{tafrishi2021darboux}. These contravariant vectors, denoted as $\bm{e} = \left[\bm{e}_1,\;\bm{e}_2,\; \bm{e}_3\right]$, on the main (fixed) considered surface frame, constitute the Darboux frame at the contact point where the angular velocity equation is described by
\begin{align}
\bm{\omega}=\delta\left(-\tau^*_g\bm{e}_1+k^*_n\bm{e}_2-k^*_g\bm{e}_3\right),
\end{align}
where $\delta$, $k^*_g$, $k^*_n$, and $\tau^*_g$ are the rolling rate, induced geodesic curvature, normal curvature, and geodesic torsion of the common contact surfaces located on that frame. Tafrishi et al. applied these Darboux-frame-based kinematics to transform Montana equations, introducing a new arclength parameter based on virtual surface theory \cite{tafrishi2021darboux}. This development led to a new kinematic model, enhancing accessibility through the introduction of new geometric parameters. However, these transformations require complex motion planning strategies since the domain is shifting from time to arclength. In the following section, we will explore how various studies, building upon the kinematic models discussed, have introduced motion planning techniques to generate a feasible path for the desired configuration of rolling contact systems.

Before delving into the motion planning problem, we present some intriguing kinematic insights derived from our generalized formulation, which we believe will be valuable to the robotics community. Montana kinematics offers simplicity and comprehensiveness, making it well-suited for tackling complex robotics problems. By excluding dynamics, including factors such as actuator influence, body inertia, and friction, the first-order equations of Montana can be directly applied to motion planning problems. We will explain later how motion planning and trajectory tracking using dynamics can be decoupled depending on the problem.
\begin{figure}[b!]
	\centering
	\includegraphics[scale=.28]{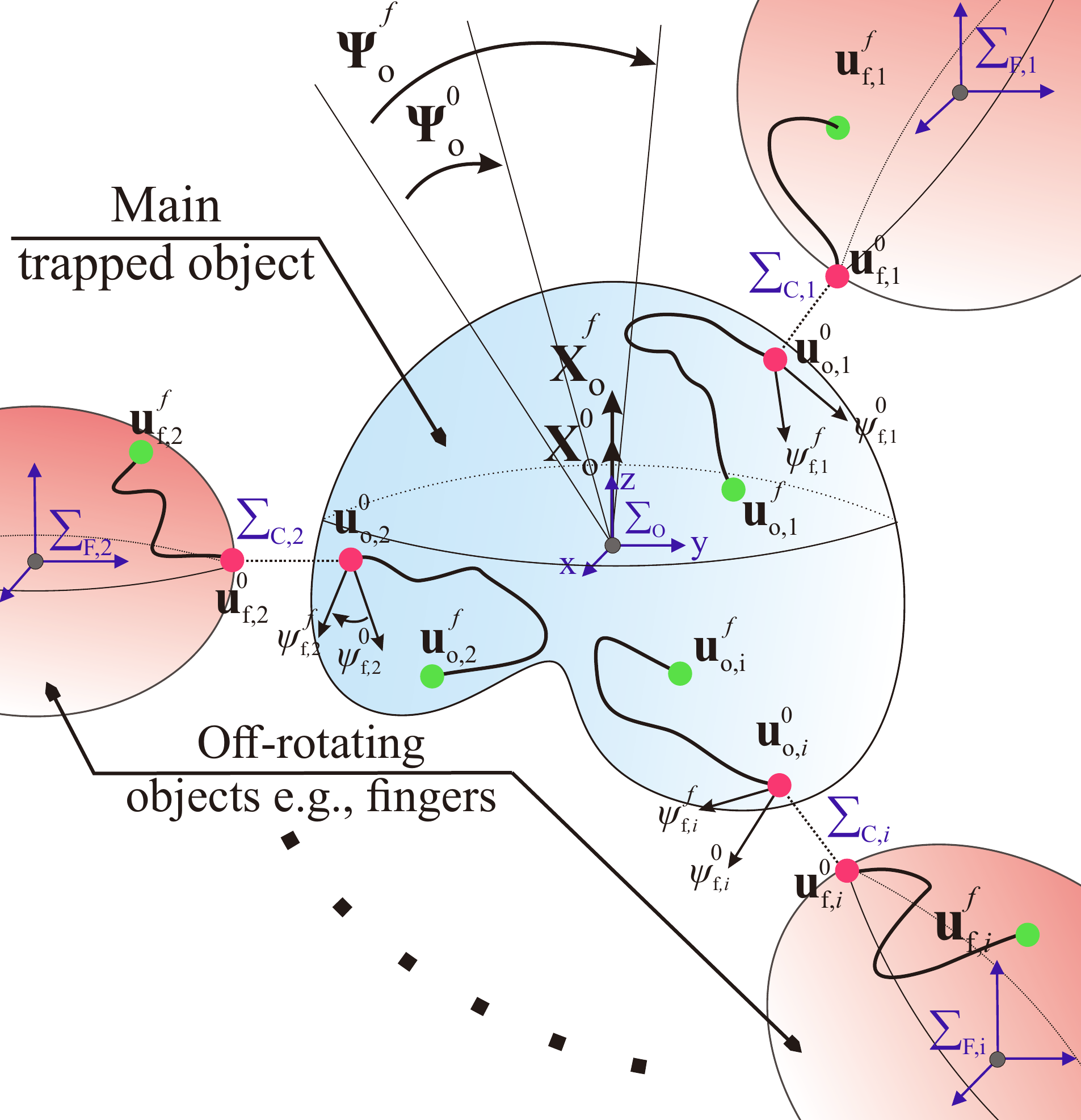}
	\caption{The path planning of $i$ number of spin-rolling convex off-rotating surfaces/fingers $U_{f,i}$ on an arbitrary surface of the main trapped object $U_o$.}
 \label{Fig:PathPlanningMultifingergraspcdr}
\end{figure} 
 
At first, consider an arbitrary $i$-th spin-rolling object $U_{f,i}$, e.g., a finger,  represented in $\Sigma_{F,i}$ frame with local coordinates $\dot{\uvec{u}}_{f,i}$, in contact with another primary object e.g., grasped/trapped object $U_{o}$, defined in $small \Sigma_{O}$ frame with local coordinates $\dot{\uvec{u}}_{o,i}$, as illustrated in Figure \ref{Fig:PathPlanningMultifingergraspcdr}. The relative spin angle between these contact coordinates is denoted as $\psi_i$. The Montana kinematics \cite{montana1988kinematics} governing the interaction between the main object surface $U_o$, with curvature properties $\{\uvec{M}_{oi}, \uvec{K}_{oi}, \uvec{T}_{oi}\}$, and the $i$-th off-rotating object surface $U_{f,i}$, with curvature properties $\{\uvec{M}_{fi}, \uvec{K}_{fi}, \uvec{T}_{fi}\}$, can be reformulated using our specific approach. This results in Equation (\ref{Eq:Montanakinemitcstransformed}), where $\bm{\omega}_{rel,i} = [\omega_{x,i},\; \omega_{y,i},\; \omega_{z,i}]^T$ and $\uvec{v}_{rel,i} = [v_{x,i},\; v_{y,i},\; v_{z,i}]^T$ represent the relative angular and linear velocities between the rolling objects. Each surface has three parameters $\{\uvec{M}, \uvec{K}, \uvec{T}\}$ that describe the geometric and curvature functions of the surfaces, which are detailed in the Appendix \ref{MultiFingerComputAppendix}. Also, the contact kinematics (\ref{Eq:Montanakinemitcstransformed}) have following matrix definitions 
\begin{equation*}
\uvec{E}_1 =\left[\begin{array}{c}
0 \\ 0 \\ 1
\end{array}\right]^T, \;\uvec{E}_2=\left[\begin{array}{ccc}
0 & 1 & 0\\
-1 & 0  & 0 
\end{array}\right] , \uvec{E}_3=\left[\begin{array}{ccc}
1 & 0 & 0\\
0 & 1  & 0 
\end{array}\right].
\end{equation*}
It is important to note that the kinematics are derived with the assumption that contact is always maintained; hence, there is no sliding in the normal direction, i.e., $v_{z,i}=0$. Note that although the spin-rolling single contact kinematics (\ref{Eq:Montanakinemitcstransformed}) finds broad application in numerous problems such as ball-plate systems and spherical robots, it represents a simplified scenario. Studies involving tasks like grasping or interactions with multi-contact robots often entail more than one object in contact with the surface. For instance, Sarkar developed a model accommodating only two contact points \cite{sarkar1997control, Sarker3Dmulticontact} that indicates the potential. 

Drawing on this motivation, we extended the model to encompass multi-contact surfaces, as depicted in Figure \ref{Fig:PathPlanningMultifingergraspcdr}. We believe that this generalization will empower the robotics community to readily apply the model to more intricate challenges, including multi-fingered grasping, specialized manipulation tasks, and research in rolling swarm robotics. In our extension of the Montana kinematics, we incorporate the orientation of the considered object. The conventional Montana kinematics primarily focuses on the local coordinates of the contacting object, often overlooking the relative orientation of the grasped or in-contact object. This aspect becomes crucial, especially with an increased number of contacts. Consequently, additional kinematic equations become imperative, and we look to derive these extended kinematics, drawing inspiration from Sarkar's foundational work for a trapped object with two rotating contacts \cite{Sarker3Dmulticontact}.

To find multi-contact kinematics of multiple bodies as Figure \ref{Fig:PathPlanningMultifingergraspcdr}, we first assume the relative rotation angular velocities on contact frames $\Sigma_{C,i}$ are defined as follows 
\begin{equation}
\left[\begin{array}{c}
\bm{\omega}_{rel,1} \nonumber \\
 \vdots \nonumber\\
\bm{\omega}_{rel,i} \nonumber\\
\uvec{v}_{rel,1} \nonumber\\
\vdots \nonumber \\
\uvec{v}_{rel,i} 
\end{array}\right] = \left[\begin{array}{c}
 \bm\omega_{f,1}-\bm\omega_{o} \nonumber \\
 \vdots \nonumber\\
 \bm\omega_{f,i}-\bm\omega_{o} \nonumber\\
\uvec{v}_{f,1} - \uvec{v}_{o}  \nonumber\\
\vdots \nonumber \\
\uvec{v}_{f,i} - \uvec{v}_{o} 
\end{array}\right],
\label{Eq:Angularrelativeveloeq}
\end{equation}
where $[\bm{\omega}_{rel,1},\dots,\bm{\omega}_{rel,i}]\in\mathbb{R}^{3(n-1)}$ and $[\uvec{v}_{rel,1},\dots,\uvec{v}_{rel,i}]\in\mathbb{R}^{3(n-1)}$ are relative velocities between the $i$ number of off-rotating objects (e.g., fingers) on $\Sigma_{F,i}$ frame with $\{\bm{\omega}_{f,i},\uvec{v}_{f,i}\}$ and main object (that all contacts are in common) $ \Sigma_{O}$ frame, and also $n$ is the number of total rotating objects and $\{\bm{\omega_o},\uvec{v}_o\}$ is the velocities for the main rotating object. We arrive at the final form of the equations that provide us with the comprehensive kinematic model for the multi-object system, incorporating each $i$-th local coordinate of the main object, off-rotating object, and the relative angle between the two surfaces, denoted as $\left(\dot{\uvec{u}}_{o,i},\dot{\uvec{u}}_{f,i}, \dot{\psi}_i\right)$. This is achieved by substituting the derived formulation with the Euler transformation in (\ref{Eq:AngularRelationBasiangularvel})-(\ref{Eq:AngularRelationBasiclinearvel}) (for detailed computation, please refer to Appendix \ref{MultiFingerComputAppendix}) which becomes
\begin{eqnarray}
\dot{\uvec{U}}=\left[\begin{array}{c}
\dot{\uvec{u}}_{o,1}\\
\dot{\uvec{u}}_{f,1}\\
\dot{\psi}_1\\
\vdots \\
\dot{\uvec{u}}_{o,i}\\
\dot{\uvec{u}}_{f,i}\\
\dot{\psi}_i\\
\end{array}\right]=\uvec{D}\dot{\uvec{q}}=\uvec{D}\left[\begin{array}{c}
\dot{\uvec{X}}_o\\
\dot{\bm{\Psi}}_o\\
\dot{\uvec{X}}_{f,1}\\
\dot{\bm{\Psi}}_{f,1}\\
\vdots \\
\dot{\uvec{X}}_{f,i}\\
\dot{\bm{\Psi}}_{f,i}\\
\end{array}\right],
\label{Eq:FinalGeneralKinematics}
\end{eqnarray}
where $\{\dot{\uvec{X}}_o,\dot{\bm{\Psi}}_o\}$ and $\{\dot{\uvec{X}}_{f,i},\dot{\bm{\Psi}}_{f,i}\}$ are defined linear and angular velocities on their surface centers $\Sigma_o$ and $\Sigma_{F,i}$ with respect to the inertial frame, respectively, and  
\begin{align}
\uvec{D}\left(\uvec{M},\uvec{K},\uvec{T}\right)=\left[\begin{array}{ccccc}
 & \uvec{D}_{f,1} & \uvec{0}  & \uvec{0} & \uvec{0}\\
  & \uvec{0} & \uvec{D}_{f,2} &\uvec{0} &\uvec{0}\\
\uvec{D}_o &   &   & \ddots  & \\
& \uvec{0} & \uvec{0} & \uvec{0} & \uvec{D}_{f,i}
\end{array}\right].
\label{Eq:Dtotalparameteroverall}
\end{align} 
Next, the parameters in (\ref{Eq:Dtotalparameteroverall}) are given by 
\begin{equation*}
\uvec{D}_o=\left[\begin{array}{cccccccc}
 ^1 \uvec{D}_{o,1}&^3 \uvec{D}_{o,1} & ^5 \uvec{D}_{o,1}& \cdots &  ^1 \uvec{D}_{o,i}&^3 \uvec{D}_{o,i} & ^5 \uvec{D}_{o,i} \\
^2 \uvec{D}_{o,1}&^4 \uvec{D}_{o,1}  & ^6 \uvec{D}_{o,1} & \cdots &  ^2 \uvec{D}_{o,i}&^4 \uvec{D}_{o,i} & ^6 \uvec{D}_{o,i}
\end{array}\right]^T
\end{equation*} 
\begin{align}
& \uvec{D}_{f,1}= \left[\begin{array}{cc}
^1 \uvec{D}_{f,1} & ^2 \uvec{D}_{f,1} \\
^3 \uvec{D}_{f,1}  & ^4 \uvec{D}_{f,1} \\
^5 \uvec{D}_{f,1} & ^6 \uvec{D}_{f,1}
\end{array}\right] \cdots \uvec{D}_{f,i}= \left[\begin{array}{cc}
^1 \uvec{D}_{f,i} & ^2 \uvec{D}_{f,i} \\
^3 \uvec{D}_{f,i}  & ^4 \uvec{D}_{f,i} \\
^5 \uvec{D}_{f,i} & ^6 \uvec{D}_{f,i}
\end{array}\right]. 
\end{align}
The details and computations of the sub-matrices $\{^j\uvec{D}_{o,i}$, $^j\uvec{D}_{f,i}\}$ are presented in (\ref{Eq:Domatricesdetails})-(\ref{Eq:Dfmatricesdetails}) in Appendix \ref{MultiFingerComputAppendix}. One can obtain the acceleration by taking the derivative of Equation (\ref{Eq:FinalGeneralKinematics}) which follows as 
\begin{equation}
\ddot{\uvec{U}}= \uvec{D}\;\ddot{\uvec{q}}+\dot{\uvec{D}}(\bm{\varGamma},\uvec{L})\;\dot{\uvec{q}}.
\end{equation}
However, this means the problem will get intricate since the second order of curvature properties should be computed which results in the second kind, and coefficients of the second fundamental \cite{sarkar1996velocity,woodruff2019second} form with Christoffel symbols $\{\bm{\varGamma},\uvec{L}\}$. Woodruff and Lynch \cite{woodruff2019second} discussed the challenges of acceleration, e.g., sign inversions in kinematic models of S-CR rotating bodies.

 \section{Path Planning} 
\label{Sec:PathPlanning}
\subsection{Singular Contact Rotation (S-CR)}
In this section, we explore various kinematic-level path-planning strategies tailored for rolling systems. The focus lies predominantly on scenarios with a singular point of contact (S-CR). In the motion planning problem, we denote the initial states (start) as $\big(\uvec{u}^0_{o},\uvec{u}^0_{f},\psi^0\big)$, while the desired states (goal) are represented as $\big(\uvec{u}^f_{o},\uvec{u}^f_{f},\psi^f\big)$. The objective is incrementally rotating the objects through infinitesimal motion steps (phases) to reach the specified goal. It's worth noting that most studies operate under the assumption of a no-sliding constraint.
\subsubsection{Singular Disk Rolling}
A rotating disk serves as a simplified scenario in S-CR kinematics with a non-holonomic constraint \cite{bloch2003nonholonomic}, involving only two rotational degrees of freedom $(\uvec{u}_{f},\psi) = (u_{f},0,\psi)$. In this setup, the disk rotates by $u_f$ with one additional angle, accounting for the spin motion $\psi$ that changes the direction of the disk, including rolling surface states $\uvec{u}_{o}$. An interesting relationship exists between simple disk rotation and cart-based systems. If we assume the inputs of rotating disks as $\bm{\omega}_{rel}=\left(\omega_x,\omega_z\right)$, they can be mapped as $\left(\omega_x,\omega_z\right) \rightarrow\left(v_x,\omega_z\right)$ between rotating wheels and a mobile robot \cite{laumond1998robot,tafrishi2022novel}. Consequently, this model can be applied to a differentially driven cart-based system, with motion planning remaining largely consistent \cite{laumond1998robot} with only three states $\left(u_o,v_o,\psi \right)$ to plan.

In the context of path planning for disk systems (e.g., carts), the initial and desired configurations of the vehicle, encompassing its position and orientation, are typically assumed to be known \cite{zheng1993recent,laumond1998robot}.  The objective is to establish a feasible trajectory for the vehicle, taking into account specific constraints, notably the presence of known obstacles. For instance, a nilpotent form of the model has been devised to facilitate feedback transformations for various systems, including cars \cite{261508}. Laumond et al. have developed a range of geometric control strategies to achieve collision-free path planning for wheeled mobile robots \cite{laumond1998robot,laumond1994motion,giordano2006nonholonomic}. As the rolling disk, the motion planning problem is also looked for moving in arbitrary surfaces with prescribing series of motion \cite{rehan2018control}.  This problem is also known as the car parking problem in different literature. Specialized planning techniques have been proposed in more challenging scenarios, such as path planning through environments with moving obstacles \cite{hsu2002randomized,lavalle2006planning,minguez2016motion}. Recent advancements in this field have taken a decisive turn towards algorithmic motion planning by integrating different sensors like LiDAR and RGB-D cameras, particularly for navigating through densely populated environments through semi-autnomous cars \cite{gasparetto2015path,paden2016survey,badue2021self}. 

While certain researches have explored the geometric configurations of disk-on-disk \cite{DiskonDiskLynch2013} and disk-on-beam \cite{DiskonBeam2019} systems, the emphasis has predominantly been on control theory aspects, addressing factors like slippage and governing rotational dynamics. Motion planning has not been the primary focus in these systems \cite{ruggiero2018nonprehensile}, as trajectory planning in these scenarios doesn't inherently simplify the problem dimensionality ($2\times1$). Thus, control problems will be more effective in direct control of rolling bodies that do not have non-holonomic constraints \cite{DiskonDiskLynch2013,tahara2010external,Ollie2025Robosoft} since states can converge directly to desired states. Insights from challenges will be discussed in the nonprehensile Section \ref{nonprehensilemaniapplicaiton} could offer valuable considerations for kinematic planners, particularly in scenarios involving slippage, such as object manipulation or throwing using the palm of the hand.
\subsubsection{Pure Rolling Motions}
The study of pure rolling in S-CR represents an extensively explored area, with relevance to numerous practical problems in robotics. Pure rolling motion refers to a scenario where the rotational object most cases convex surfaces like a sphere exhibits only two rotational motions, expressed as {\small $\bm{\omega}_{rel}=\left(\omega_x,\omega_y\right)$}, without any spinning/twisting rotation. These investigations primarily operate under the no-slippage constraint. Researchers have endeavored to devise various strategies or sequences of motions to guide the rotating object, such as a sphere, to specific desired states  $\big(\uvec{u}^f_{o},\uvec{u}^f_{f},\psi^f\big)$.
\begin{figure}[t!]
	\centering
	\includegraphics[width=5 in]{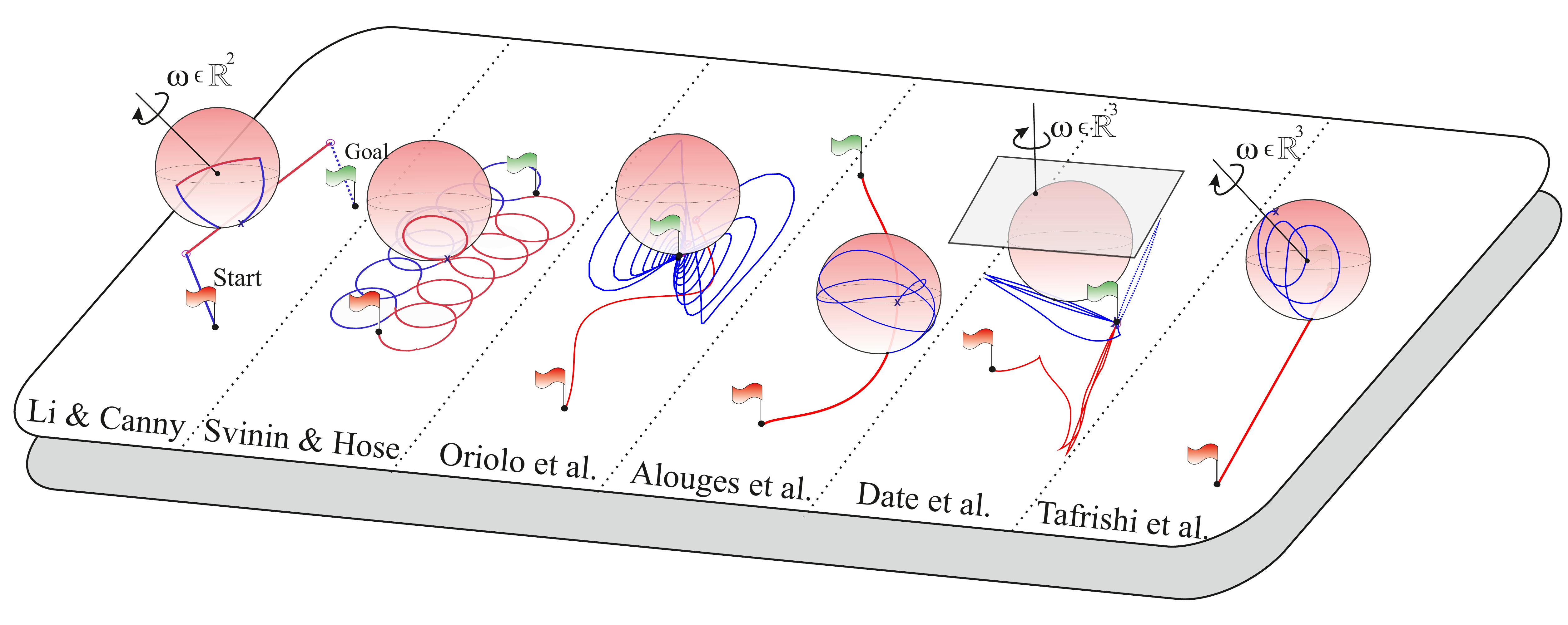}
	\caption{Simulation results of some well-known motion planning strategies for the rolling contact kinematics.}
\label{Fig:ResultsPhaseShift}
\end{figure} 

The sphere is mostly considered a geometrical form $U_f$ that is easy to study in motion planning problems due to its symmetrical form and clear presentation of projected trajectories. The task of finding a trajectory that lets a rotating object like a sphere reach the desired configuration on a fixed surface like a plane is famous for a \textit{ball-plate} problem, which has historically opened many interesting discussions in the mathematical community. The planning for pure rolling motion with nonholonomic constraints can be classified into two main categories, some of whose solutions are depicted in Figure \ref{Fig:ResultsPhaseShift}: feedforward and feedback methods.

Feedforward planning can be further divided into two distinct approaches. The first approach is grounded in optimal control theory \cite{arthurs_walsh_1986,jurdjevic1993geometry,sachkov2010,mashtakov2011}, providing a theoretical framework for determining optimal trajectories for moving a convex surface like a sphere. In early studies, Arthurs et al. \cite{arthurs1986hammersley} demonstrated that optimal solutions can be expressed in terms of elliptical integrals with a boundary value problem. Additionally, Jurdjevic \cite{jurdjevic1993geometry} showed that such optimal curves can be characterized as Euler elasticae, which bring a pure rolling sphere to specific states. This research direction, describing continuous and discrete symmetries and deriving equations on Maxwell points (points reachable by multiple extremals at the same time), was further advanced by Sachkov \cite{sachkov2010,mashtakov2011}. Following these studies, a simulation study of a rolling sphere on an inhomogeneous surface identified similar extremals for certain desired states using optimal conditions developed through Hamiltonian systems \cite{mashtakov2021extremal}. The latter feedback approach leverages geometric phases \cite{murray1994mathematical,li1990motion,levi1993geometric}, which utilize closed paths of control inputs to induce changes in contact coordinates. Li and Canny \cite{li1990motion} proposed a three-step algorithm for motion planning under consideration of Lie groups to converge the sphere to certain states. While optimal control theory shows promise in theory, its practical computational feasibility remains untested. On the other hand, geometric phase approaches have been refined over time, streamlining the number of maneuvers required for effective planning \cite{mukherjee2002feedback}, inspired from Ref. \cite{li1990motion}. An alternative algorithm, based on state equation transformation and piecewise constant inputs, was also proposed for solving nonlinear algebraic equations \cite{marigo2000rolling}. Additionally, there have been attempts to extend the Li and Canny method \cite{li1990motion} to sphere-on-sphere problems with certain adjustments in the number of steps \cite{rehan2018global}. Alouges et al. demonstrated the use of numerical continuation methods to generate algorithmic open-loop motion plans for various convex surfaces, such as spheres or elliptical geometries, rolling on a plane \cite{alouges2010motion}, considering slippage constraints.

Other pioneering study addressing motion planning under contact area constraints is found in the realm of rolling-based manipulation \cite{harada2002rolling}. This approach constructs multiple spherical triangles on object surfaces, the size of which is determined by considerations of contact stability \cite{montana1992contact}. Another noteworthy study \cite{bicchi2002dexterous} introduces a local-local algorithm, which can be adapted to deal with limited contact areas. It employs geodesic quadrilaterals in its nontrivial maneuver, the size of which depends on the workspace limitations. Similarly, a somewhat related algorithm proposes iterating trapezoidal paths on the sphere \cite{nakashima2005control}. These algorithms produce piecewise smooth trajectories, but they may require stops at polygon vertices to ensure movements remain impact-free. To address this, Svinin and Hose proposed by replacing polygons with smoother curves $C^{\infty}$ emerges as a more advantageous strategy that is composed of circles and generalized Viviani curves as feedback-based solution \cite{svinin2008motion} to arrive the sphere to desired states as Figure \ref{Fig:ResultsPhaseShift}. As an alternative, Oriolo et al. formulated an iterative steering method to stabilize the non-differentially flat ball-plate system \cite{Oriolo2005Feedback} with feedback formulations. However, the solutions exhibited significant fluctuations as trajectories converged to the desired states, making them challenging to implement in dynamical systems \cite{morinaga2014motion}. Additionally, singularities arose in different regions of the spherical manifold due to the localized nature of the solutions. 

It's worth noting that many of the algorithms proposed so far are primarily tailored for spherical shapes. This limitation poses challenges when dealing with differently shaped convex rolling objects or surfaces with varying curvatures which some geometric \cite{alouges2010motion,Tafrishi2019} or phase shift based \cite{planningli1990,mukherjee2002motion,svinin2008motion,kilin2015spherical,bai2018dynamic} methods might mitigate but have not well studied before. Additionally, the motion planning problem in pure rolling, constrained by spinning, requires extensive utilization of plane states to orient the sphere as desired. For geometric phase-based methods, when rolling is dynamically constrained, the property of geodesic-to-geodesic mapping does not necessarily hold true \cite{svinin2013dynamic}. Consequently, the aforementioned algorithms cannot be directly applied without significant modifications, as the trajectories they generate may not be dynamically realizable in general.

\subsubsection{Spin-Rolling Motions}
Analyzing spin-rolling motion with  $\bm{\omega}_{rel}=\left(\omega_x,\omega_y,\omega_z \right)$ inputs using conventional ball-plate system kinematics is possible \cite{jurdjevic1993geometry,marigo2000rolling}, but directly spinning the ball through the rotation of the upper sandwiched plane is not applicable \cite{jurdjevic1993geometry,tafrishi2021darboux}. Various kinematic parametrizations for rolling contact have been explored \cite{sankar1996velocity,Montana1988,cui2010darboux,bizyaev2019different,woodruff2019second,tafrishi2021darboux} to realise the spin-rolling motions.

The first planning strategy that incorporated indirect spinning to control the ball-plate system was proposed by Date et al. \cite{date2004simultaneous}, as illustrated in Figure \ref{Fig:ResultsPhaseShift}. Their motion planning algorithm involved iteratively shifting the coordinate of the actuating plane with respect to different reference frames, akin to a rotational virtual center. However, time scaling combined with coordinate transformation in the kinematic model could lead the system to an uncontrollable state \cite{Oriolo2005Feedback}.

In the domain of pure rolling, a range of feed-forward \cite{jurdjevic1993geometry,das2004exponential,alouges2010motion,svinin2008motion} and feedback control \cite{date2004simultaneous,Oriolo2005Feedback,mukherjee2002feedback,woodruff2020motion} methods have been developed. Nonetheless, due to the influence of spin angle changes on sphere orientation, conventional planning approaches, such as those based on geometric phase shifting \cite{planningli1990,mukherjee2002motion,svinin2008motion,kilin2015spherical,bai2018dynamic}, do not naturally adapt for spin-rolling motion planning. Beschastnyi \cite{beschatnyi2014optimal} worked on the optimality problem of the spin-rolling sphere. Extremal trajectories were parameterized, and their cut times were estimated for optimality. It was suggested that the Maxwell time could be determined while the sphere follows a straight path to achieve the optimal solution. However, this research did not address the control problem for arbitrary desired states. Recently, there have been attempts to create motion planning solutions using Darboux-frame transformation \cite{tafrishi2021darboux} that reduces the complexity. The geometric motion planning was designed to create multiple admissible smooth trajectories under optimal shortest trajectory on the plane \cite{tafrishi2023geometric}.

It is crucial to emphasize that feedback-based planning algorithms \cite{date2004simultaneous,Oriolo2005Feedback, das2004exponential, mukherjee2002feedback} generate piecewise-smooth trajectories. Significantly, some methodologies eliminate the need for decomposing the planning strategy into motion steps, a characteristic found in geometric phase shift methods \cite{planningli1990,mukherjee2002feedback,date2004simultaneous,morinaga2014motion,tafrishi2023geometric}. Additionally, Tafrishi et al. introduced a differential geometry-based method using arclength-based virtual surface inputs that offers the advantage of shaping the resulting angular velocities of the spin-rolling sphere to achieve different desired convergence rates, enabling tuning of various convergence trajectories, and allowing arbitrary optimal paths on a plane \cite{tafrishi2023geometric}. Despite these advantages, methods based on search algorithms \cite{date2004simultaneous,alouges2010motion,tafrishi2023geometric} still face challenges in achieving real-time robust solutions due to offline computations.

Several other significant challenges exist in motion planning, notably the absence of studies encompassing a generalized motion planner for arbitrary geometries such as sphere-on-sphere interactions or more time variable geometries which is predominant in smooth and deformable surfaces. The complexity arises from the need to verify controllability, necessitating continuous checks of LARK \cite{wang2011note,lian1994controllability} to ascertain if certain geometries are accessible and controllable based on inputs or reduced sub-systems. While some geometric methods can be extended to complex or arbitrary geometries \cite{svinin2008motion,alouges2010motion,li1990motion,tafrishi2023geometric}, this limitation restricts the applicability of methods to a great extent. Another issue is the presence of slippage in real applications, which fundamentally alters the kinematics model and can result in a drift system with potentially unbounded states. Integrating this factor into the problem or decoupling it from the planner remains an ongoing area of research in the field of rolling systems.

\subsection{Multiple Contact Rotation (M-CR)}
In this section, we begin by describing the general definition of motion planning for multiple contact rotation (M-CR) of objects. We trace its origins back to the field of grasping problems and then elaborate on its broader applications, including scenarios involving a swarm of rolling robots. Finally, we present potential hypothetical solutions for extending single-contact rotation (S-CR) planning approaches to address M-CR challenges.

As the path planning problem, we consider the main object $U_o$ as shown in Figure \ref{Fig:PathPlanningMultifingergraspcdr} always trapped within the $i$ number of spin-rolling objects. The initial and desired states of all objects are given. The goal is to find the admissible smooth paths to converge the local desired coordinate of each contact coordinate $\{\uvec{u}^f_i,\uvec{u}^f_{o},\psi^f_i\}$ while the rotating main objects state $\{\uvec{X}^f_o,\bm{\Psi}^f_o\}$ are reached at the same time. It is important to note that we ignore the states $\{\uvec{X}^f_{f,i},\bm{\Psi}^f_{f,i}\}$ of the off-rotating objects $U_{f,i}$ are not considered since these states change based on the path planning solutions and the off-rotating objects. The off-rotating objects/surfaces can be looked at as fingers in grasping simple scenario states that do not pose any importance.

While there has been extensive research in the dexterous manipulation and grasping community regarding the control of stability for multi-contact objects \cite{ozawa2017grasp}, the field of multi-contact motion planning remains relatively underexplored in the literature for the case the fingertips are not fixed on the contact. Some researchers refer to this kind of problem as \textit{nonprehensile manipulation} \cite{ruggiero2018nonprehensile}.  In a brief review of dexterous manipulation, Okamura identified remaining challenges related to rolling and sliding, particularly in motion planning the points of contact \cite{okamura2000overview} in a brief manner. In this study, we do not look for high-level discrete motion analyses for object motion, which most recent studies covered on \cite{ozawa2017grasp,yu2022dexterous,pfanne2022hand}. In our problem, we consider the cases in which rotating objects (can be fingertips) are always in contact with the main object (grasped object). Then, the contact motion planning can be examined in two potential scenarios of grasping: a) continuous grasping with agile fingertips, and b) regrasping with fixed fingertips. In the case of agile fingertips, assuming continuous contact with the grasped object ($v_z=0$) during a precision grasp, Kiss et al. \cite{kiss2002motion} employed three independent planes to manipulate the sphere at the kinematic level, controlling the relative angles of the sphere without considering the desired moving plane configuration \cite{kondo2008recognition}. Regrasping and finger-gaiting operations involve a combination of continuous dynamic systems (covering manipulation kinematics, dynamics, and factors like gravity and slipping) and discrete-event systems (such as contact or detachment events). This necessitates the analysis and control of \textit{hybrid} systems, which are a blend of event-driven and time-driven processes. The stability analysis and formal verification in the context of automata theory present a complex, open problem for the fields of computer science and automatic control which we do not focus on here. Initial studies in robotics applications have provided some insights \cite{rus1999hand,kondo2008recognition,bicchi2000hands}, but substantial challenges persist. Additionally, optimizing execution times for re-grasping plans and establishing their robustness, particularly for complex 3-D objects, remain key, unresolved issues in this domain. High-level studies on re-grasping through learning methods have been extensively explored in the literature and are rapidly expanding \cite{xu2010sampling,shi2017dynamic,khadivar2023adaptive}. However, there is a lack of in-depth technical development of low-level motion planning in nonprehensile manipulation to realize contact point motion tracking while there are discrete actions. This gap may be attributed to the significant challenges associated with tracking the contact point using certain sensors. This topic will be discussed in more detail in the upcoming section.   

Continuous contact systems in M-CR can encompass applications in rotational reconfigurable robots \cite{moubarak2012modular,seo2019modular,zong2022kinematics}, grasping mechanisms with agile fingertips \cite{tahara2012externally,ozawa2017grasp,yuan2020design,yuan2022robot}, or multiple swarms of particles \cite{sun2014liquid} in 3D space. The assumption of continuous contact is generally guaranteed, allowing us to decompose the problem into multiple layers. Here, we like to discuss the kinematic aspects of the problem. The subsequent section will incorporate dynamics and mechanical properties into the analysis. To the best of our knowledge, there hasn't been any study that connects the S-CR planning strategies to M-CR and simplifies the problem for better understanding. Therefore, we propose a different hypothesis to approach motion planning by simplifying it down to an S-CR problem, as follows:

 \begin{figure}[t!]
	\centering
	\includegraphics[scale=.3]{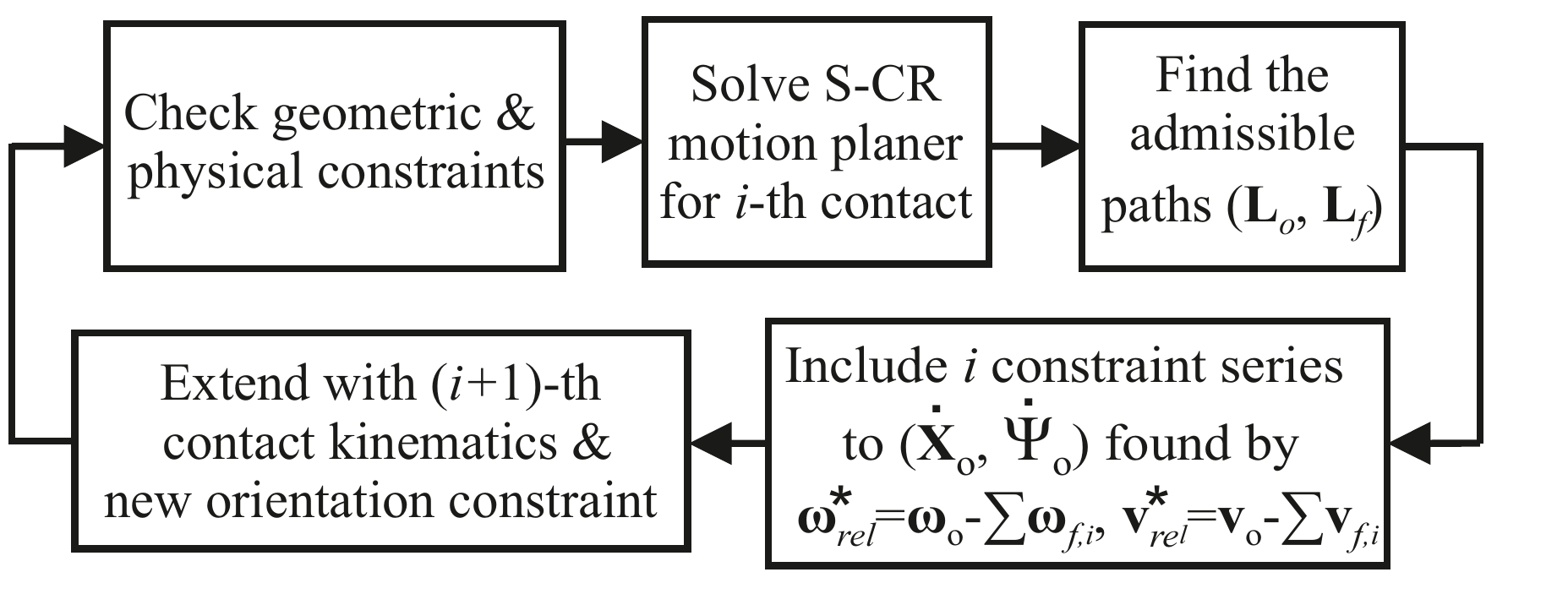}
	\caption{An incremental path planning problem strategy for M-CR.}
	\label{Fig:Motion_Planning_MCR}
\end{figure} 

\subsubsection{Incremental Planning} This approach initially addresses the motion planning problem for one of the rotating objects (fingers) on the main object. Once a solution is obtained for this initial finger, the process is repeated sequentially for the remaining fingers, as depicted in Figure \ref{Fig:Motion_Planning_MCR}. This entails first considering the planning problem within the context of single contact rotation (S-CR), accounting for the specific area of contact on both surfaces. Subsequently, after determining admissible paths on both the rotating objects $U_{f,i}$ and the main object $U_o$, the resulting relative angular velocity serves as a constraint for the motion planning of the next rotating object (finger). This iterative process involves expanding the kinematics to include one additional rotating object. The angular velocity constraints affect the motion planning of the next rotating object (finger) relative to the main object $(\bm{\omega}^*_{rel}=\bm{\omega}_o -\sum \bm{\omega}_{f,i}$ and $\uvec{v}^*_{rel}=\uvec{v}_o -\sum \uvec{v}_{f,i})$. At every stage, an additional layer of complexity arises, necessitating the implementation of obstacle avoidance measures to prevent collisions with the trajectories traversed by other rotating objects (fingers). For example, this could happen due to constraints on fingers and fingertips in dexterous manipulation. 

In the state of solving the motion planner by S-CR approaches, the feedforward strategies might not be feasible since there will be constant drift due to internally added velocity changes. Thus, the feedback methods \cite{date2004simultaneous,Oriolo2005Feedback,mukherjee2002feedback,woodruff2020motion,tafrishi2023geometric} could be the most suitable to solve the local motion planner. The phase shift strategies like those proposed by Svinin and Hosoe \cite{svinin2008motion} might also be applicable, but a new kinematics formula with a drift term should be obtained to compensate for the angular velocity deviations. Trajectory tracking techniques, coupled with incremental planning approaches, can further enhance the feasibility of rolling motion by breaking down complex paths into locally solvable segments, addressing dynamic and geometric constraints iteratively. From a control perspective, as an example superposition principle under passivity \cite{arimoto2000principle} could stabilize the system dynamics, avoiding the complexity of inverse dynamics by sequentially addressing the motions of each off-rotating (finger) body.

\subsubsection{Mirrored Motion Planning} 
\begin{figure}[t!]
	\centering
	\includegraphics[scale=.32]{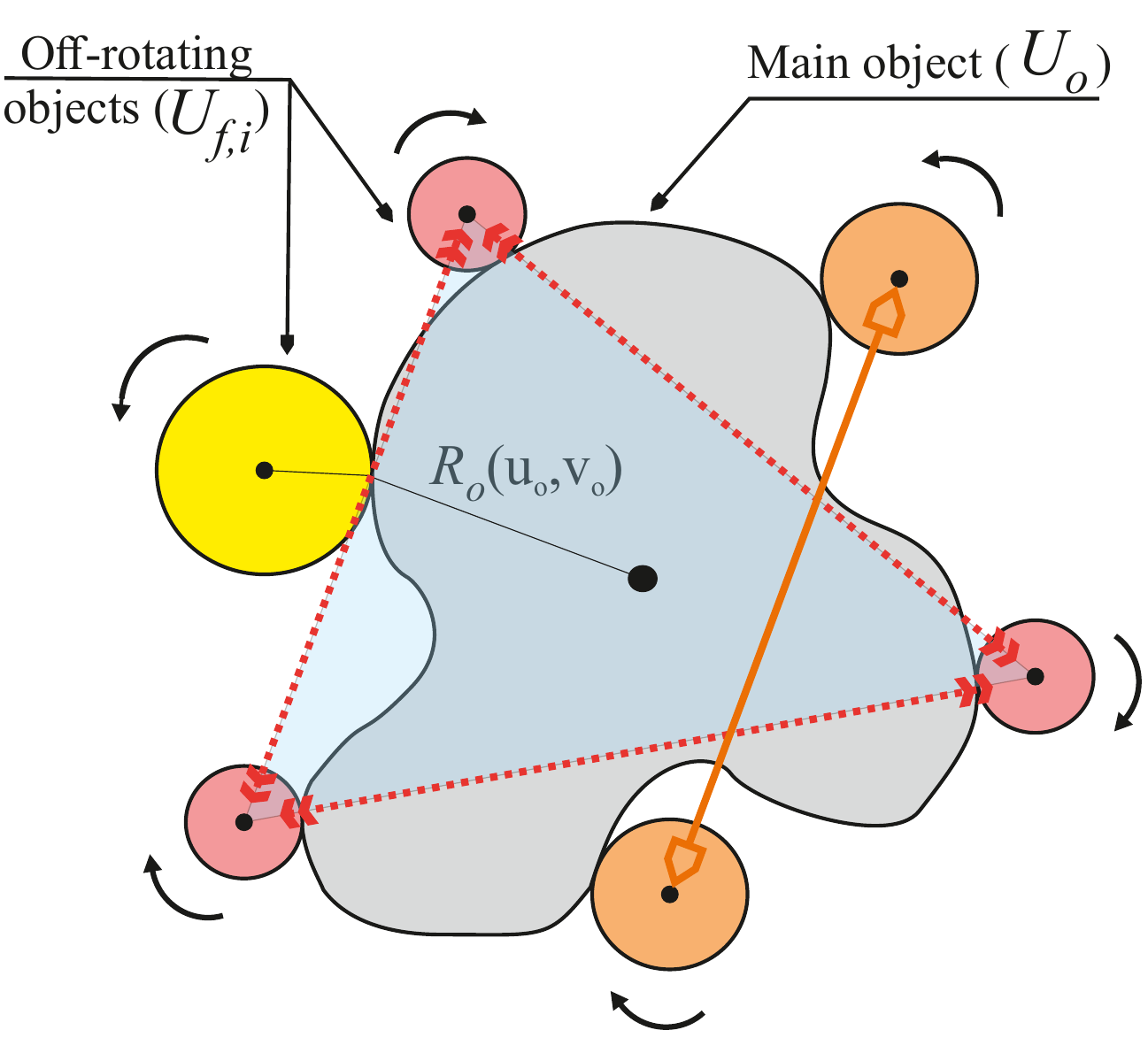}
	\caption{An example scenario for six rotating objects with arbitrary main surface objects. Note that for simplicity the rotating objects are assumed simple disks with random radii.}
	\label{Fig:Rotation_Miroring}
\end{figure}

Applying mirrored motion planning involves pairing fingers in opposite locations and solving them with mirrored trajectories. This approach simplifies the problem, as depicted in Figure \ref{Fig:Rotation_Miroring}. The key idea is to reduce the motion planning dimension by disregarding states of some rotating objects, shown in orange and red in Figure \ref{Fig:Rotation_Miroring}. The mirrored rotating objects contribute relative velocity to achieve desired states for the primary rotating object, and their trajectories are mirrored and projected onto the surface curvature on contact. This entails that to compute an exact relative velocity for each mirrored rotating object, their rotation should follow the same motion pattern as the primary rotating object, but with adjustments according to curvature changes. After determining the trajectories for the primary rotating object, the relative velocities can be computed by inverting the contact kinematics and then solving for the local rotational states of mirrored objects $(\dot{\uvec{X}}_{f,i},\dot{\bm{\Psi}}_{f,i})$. This procedure can be extended to mirrored objects. This reduction in complexity enables the application of \textit{incremental motion planning}, focusing on a limited number of rotating objects (fingers), with any S-CR approach serving as the local planner. From a control perspective, the force closure of mirrored rotating objects becomes more tractable, as the planner maintains a consistent relative distance between grouped off-rotating objects—a principle extensively studied in grasping problems \cite{nguyen1988constructing,ozawa2017grasp}.

\subsubsection{Sub-System Planning} In certain scenarios, some kinematic local states may be deemed inconsequential and can be disregarded. For instance, in grasping scenarios, the states of the continuously rotating objects e.g., rolling fingertips or swarm particles may not require specific angular convergence. The primary objective is often to achieve convergence for the main object's three position and rotation states by leveraging the fingertips/off-rotating of the grasped/trapped object. A similar approach was employed by Kiss et al. \cite{kiss2002motion}, where the fingertips of the grasped object were treated as planes, facilitating the transformation of the confined sphere into the desired states. Consequently, rotating objects that are not essential for convergence can be decoupled from the problem statement.

Here, we aim to derive a generalized formulation for decoupling the states for planning purposes. Initially, we can reformulate Equation (\ref{Eq:Montanakinemitcstransformed}) for the $i$-th contact model as follows:
\begin{align}
\left[\begin{array}{c}
\dot{\uvec{u}}_{o,i}\\
\dot{\uvec{u}}_{f,i}\\
\psi_i
\end{array}\right]&=
\left[\begin{array}{cc} \uvec{g}_{o,i} & \uvec{0}  \\
\uvec{0} & \uvec{g}_{f,i} \\
\uvec{g}_{\psi o ,i}& \uvec{g}_{\psi f ,i} 
\end{array}\right]\bm{\omega}_{xy,i}+\left[\begin{array}{c} \uvec{0} \\
\uvec{0}  \\
-\omega_{z,i}
\end{array}\right] + \left[\begin{array}{ccc} \uvec{f}_{o,i} & \uvec{0} & \uvec{0} \\
\uvec{0} & \uvec{f}_{f,i} & \uvec{0}\\
\uvec{f}_{\psi o ,i}& \uvec{f}_{\psi f ,i} &  \uvec{0}
\end{array}\right] \uvec{v}_{xy,i}
\end{align}
where $\bm{\omega}_{xy,i}=(\omega_{x,i},\omega_{y,i})$ and $\uvec{v}_{xy,i}=(V_{x,i},V_{y,i})$.
Next step, we obtain the strictly triangular form using the input transformation as $\bm{\omega_}{xy,i}=\uvec{g}^{-1}_{o,i}\;\dot{\uvec{u}}_{o,i}$. After the state transformation and substituting it back to (\ref{Eq:Montanakinemitcstransformed}) will result in
\begin{align}
&\left[\begin{array}{c}
\dot{\uvec{u}}_{o,i}\\
\dot{\uvec{u}}_{f,i}\\
\psi_i
\end{array}\right]=
\left[\begin{array}{ccc} \uvec{I}_{2\times2} \uvec{E}_2 & \uvec{0} & \uvec{0} \\
\uvec{0} & \uvec{g}_{f,i}\uvec{g}^{-1}_{o,i} \uvec{E}_2 & \uvec{0}\\
\uvec{g}_{\psi o ,i}\uvec{g}^{-1}_{o,i} \uvec{E}_2& \uvec{g}_{\psi f ,i} \uvec{g}^{-1}_{o,i}\uvec{E}_2 &  -\uvec{E}_1 
\end{array}\right]\bm{w}_{rel,i}+ \left[\begin{array}{ccc} \uvec{0} & \uvec{0} & \uvec{0} \\
\uvec{0} &\uvec{f}'_{f,i} & \uvec{0} \\
\uvec{f}'_{\psi o ,i} & \uvec{f}'_{\psi f ,i} &  \uvec{0}
\end{array}\right] \uvec{v}_{rel, i}
\end{align}  
where $\bm{w}_{rel,i}=[w_{x,i},\;w_{y,i},\;\omega_{z,i}]^T$ which $(w_{x,i},\;w_{y,i})$ are the new transformed pure rolling angular velocity inputs and
\begin{align}
    &\uvec{f}'_{f,i}= \left(\uvec{f}_{f,i}-\uvec{g}_{f,i}\uvec{g}^{-1}_{o,i}\uvec{f}_{o,i}\right) \uvec{E}_3, \nonumber\\
    &\uvec{f}'_{\psi o,i}= \left(\uvec{f}_{\psi o ,i}-\uvec{g}_{\psi o ,i}\uvec{g}^{-1}_{o,i}\uvec{f}_{o,i} \right)\uvec{E}_3, \nonumber\\
    &\uvec{f}'_{\psi f,i}=\left( \uvec{f}_{\psi f ,i}-\uvec{g}_{\psi f ,i} \uvec{g}^{-1}_{o,i}\uvec{f}_{o,i} \right) \uvec{E}_3.
\end{align}
Now, the generalized multi-contact model also gets new model by using same computation ( shown in Appendix \ref{MultiFingerComputAppendix}) can have 
\begin{eqnarray}
\label{Eq:Thelocalmatrixtransformation}
\dot{\uvec{U}}_o&=\left[\begin{array}{c}
\dot{\uvec{u}}_{o,1}\\
\vdots \\
\dot{\uvec{u}}_{o,i}\\
\end{array}\right]=\uvec{D}^1_{o}\left[\begin{array}{c}
\dot{\uvec{X}}_o\\
\dot{\bm{\Psi}}_o\\
\end{array}\right]+\uvec{D}^2_o\left[\begin{array}{c}
\dot{\uvec{X}}_{f,1}\\
\dot{\bm{\Psi}}_{f,1}\\
\vdots \\
\dot{\uvec{X}}_{f,i}\\
\dot{\bm{\Psi}}_{f,i}\\
\end{array}\right]\\
\left[\begin{array}{c}
\dot{\uvec{U}}_f\\
\dot{\uvec{U}}_\psi\\
\end{array}\right]&=\left[\begin{array}{c}
\dot{\uvec{u}}_{f,1}\\
\vdots \\
\dot{\uvec{u}}_{f,i}\\
\dot{\psi}_1\\
\vdots \\
\dot{\psi}_i\\
\end{array}\right]=\uvec{D}^1_{f}\left[\begin{array}{c}
\dot{\uvec{X}}_o\\
\dot{\bm{\Psi}}_o\\
\end{array}\right]+\uvec{D}^2_f\left[\begin{array}{c}
\dot{\uvec{X}}_{f,1}\\
\dot{\bm{\Psi}}_{f,1}\\
\vdots \\
\dot{\uvec{X}}_{f,i}\\
\dot{\bm{\Psi}}_{f,i}\\
\end{array}\right].\nonumber\\
\label{Eq:FinalGeneralKiReduced}
\end{eqnarray}
Local coordinates on main rotating object $\dot{\uvec{U}}_o$ (\ref{Eq:Thelocalmatrixtransformation}) has direct control inputs as the new transformed states with linear relation (differentially flat) from both off-rotational $\{\dot{\uvec{X}}_{f,1},\dot{\bm{\Psi}}_{f,1}\cdots\dot{\uvec{X}}_{f,i},\dot{\bm{\Psi}}_{f,i}\}$ and main object $\{\dot{\uvec{X}}_o,\dot{\bm{\Psi}}_o\}$ inputs. This means the object's local states can be converged easily with any conventional planner, however, the remaining states will get a dependent behaviour based on trajectories on the main object. This problem has been looked at from phase shifting strategies \cite{Oriolo2005Feedback,svinin2008motion,tafrishi2023geometric} where different methods are proposed (for S-CR) which can be extended for M-CR problem. 

\section{Robotic Applications}
\label{Sec:RoboticApplications}
This section aims to bridge the gap between path planning strategies and various applications, including existing mechanisms such as rolling fingertips, robotic systems like spherical robots, ballbots, reconfigurable robots, as well as micro/nano-particle manipulation and nonprehensile manipulation problems. The discussion will focus on different actuation types, robots, and techniques pertinent to kinematics, which are briefly presented in Table \ref{Tab:expcase}. The aim is to showcase the potential application of proposed planning strategies, incorporating new challenges with respect to the type of contact, actuation principles, and sensory feedback. In Table \ref{Tab:expcase}, the friction at the contact point is simplified to three levels from low to high. The low level indicates negligible Coulomb friction and a strong existence of slippage where contacted surfaces (at least one) are smooth. High friction indicates a rough surface model with a high value of friction coefficient from Coulomb friction (no-slippage), and the area of contact could be more than a single friction cone model \cite{fakhari2019modeling,romeo2020methods,boucly2007modeling}. This distribution also aligns with the presented graph in Figure \ref{Fig:TheFingertipManipulationPathPlanningStatement}.
\begin{table}[t!]
\renewcommand{\arraystretch}{1}
	\caption{Existing and potential applications of rolling contacts concerning motion planning.}
	\label{Tab:expcase}
	\centering
		\begin{tabular}{ p{1.7cm}p{2.1cm} p{.8cm}p{1.2 cm}p{1.3cm}p{1.8cm}p{1.4cm}p{1.3cm} }
			\hline\hline
			Systems & Driving mechanism & Contact & Degree  of rotation & Friction & Actuator/Actor & Contact sensor & Planning options  \\
			\hline
			  Spherical robot& Torque-Reaction& 1 & 2 & High & Reaction wheels & IMU /LiDar /tactile$^\diamondsuit$ &  S-CR\\
             & Conserv. of angular momentum & 1 & 2-3 & Low  &  Inertia disks/ rotating mass & $-$ & S-CR \cite{morinaga2014motion}\\
             & Mass-imbalance& 1 & 2-3 & Medium  &  {\footnotesize Pendulum actuated/ rotating mass}& $-$ & S-CR \\
             \hline
             Ball-bot & Omni-wheel motors& $3^*<$ & 1 & High & External omniwheels& IMU$^\diamondsuit$ & S-CR/M-CR \cite{date2004simultaneous}\\  
			\hline
            Dexterous manipulation  & Fixed fingertip&$2^{\ast \bullet}< $ & 1 & High & Finger joints& Tactile/ vision\cite{navarro2021proximity} & M-CR \\
             (hand) & Rolling fingertip&$2<$ & 1-3 & Low-Medium &Internal actuator& $-$\cite{yuan2023tactile} & M-CR  \cite{li1990motion,svinin2008motion,tafrishi2021darboux}\\
            \hline
            Nonprehensile manipulation & Juggling&$2^\bullet<$ & 1-3 & High & Palm hand & Vision \cite{kober2012playing} & S-CR/ Hybrid\\
              & Catching/ Throughing&$2^\dagger<$ & 1-2 & Medium  
            & Hand/gripper & $-$\cite{batz2010dynamic} & S-CR \\
              & Bating&$1$ & 1-3 & High & Bat & $-$ & S-CR \\
             & Balancing&$2<$ & 1-3 & Low- Medium & Plate/disk platform & $-$ \cite{DiskonDiskLynch2013} & M-CR \\
			   \hline
            Legged robot's foot (each leg) & Knee and hip joints &$1^\ast$ & 1-2 & Medium-High & Passive foot& Tactile$^\diamondsuit$ \cite{chuah2019bi} & S-CR \\ 
            \hline
            Reconfigurable rolling robots  & External driver&$1^{* \dagger}<$ & 1-2 & Medium & Gear/locker & IMU$^\diamondsuit$ & M-CR \\ 
              & Internal driver&$1<$ & 1-2 & Medium & Magnetic & Magnetom.$^\diamondsuit$& M-CR \\ 
            \hline
             Micro-/nano-particle manipulation & External needle& $2^\dagger$ & 1-3 & Low & Magnetic, fluid or needle & Microscope & S-CR/M-CR \\
			\hline\hline
	\end{tabular} \newline
  $^\ast:$ Constrained area of contact, $^\bullet:$ Discrete actions required, $^\dagger:$ Dynamic constraints, \\ $^\diamondsuit:$ Proposed solutions 
\end{table}
\subsection{Ballbots \& Spherical Robots}
\begin{figure}[t!]
	\centering
	\includegraphics[scale=.35]{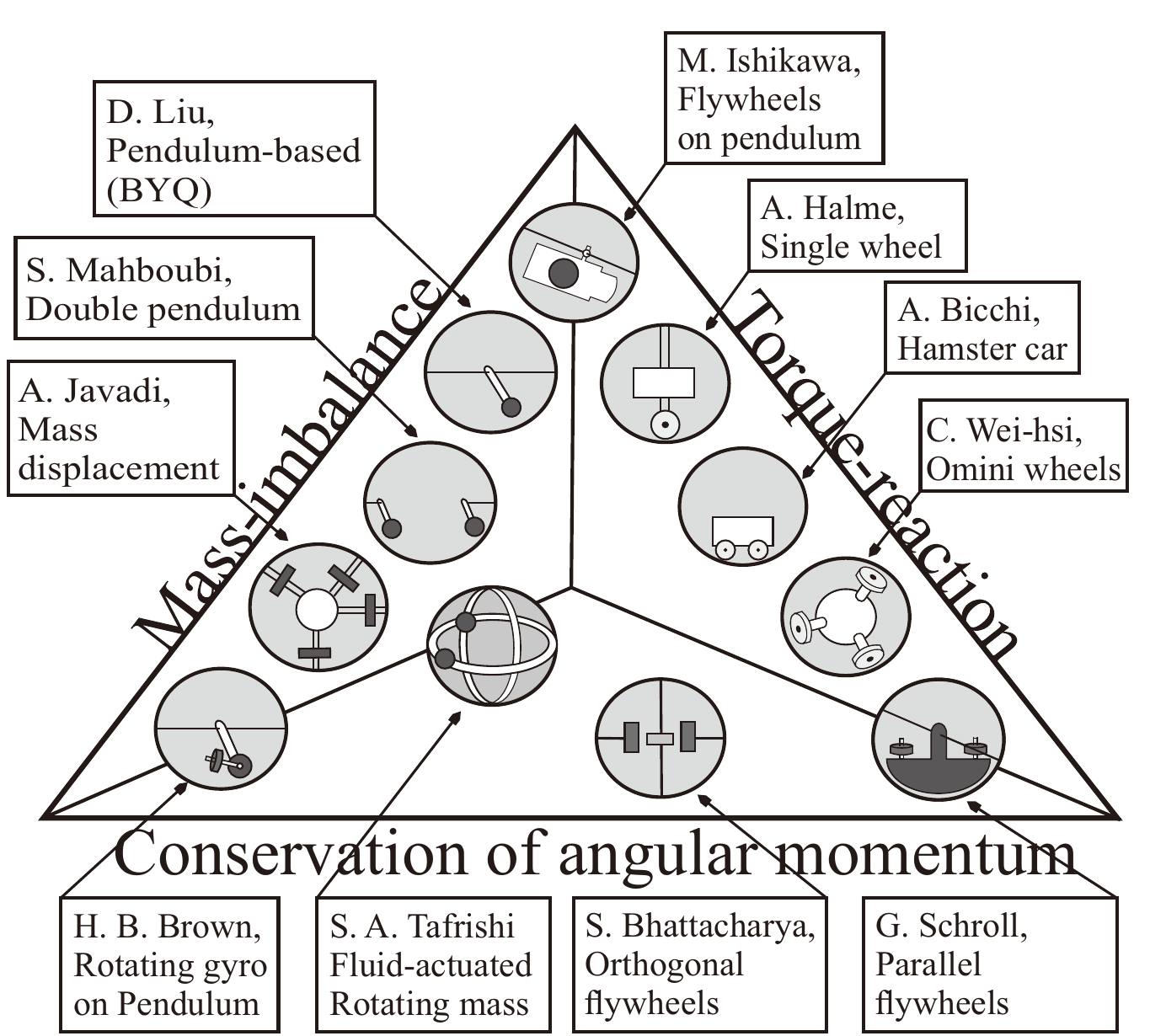}
	\caption{Different internal actuation methods for rolling robots: Torque-reaction \cite{ishikawa2011volvot,halme1996motion,Bicchi1997,chen2012design}, conservation of angular momentum \cite{schroll2008design,bhattacharya2000spherical,brown1996single}, mass-imbalance \cite{Tafrishi2019,javadi2004introducing,liu2008stabilization,mahboubi2013design}}
\label{Fig:Pastrobots}
\end{figure}
Spherical bodies in rolling motion are categorized into two types: Spherical shells equipped with internal actuators, recognized as \textit{spherical robots}, and Spherical ball working as actuators with attached external platforms, known as \textit{Ballbots}.

In recent studies, spherical robots have garnered significant interest from the community \cite{diouf2024spherical}. The actuation types for these robots are generally categorized into three types \cite{Tafrishi2019,robotics1010003,Minggang2023}, as shown in Figure \ref{Fig:Pastrobots}: torque-reaction, mass-imbalance, and conservation of angular momentum. In torque-reaction propulsion, rolling wheels create reactive torque, and various designs, such as single-direction turning wheels by Bicchi \cite{Bicchi1997} and Omni wheel platforms, have been proposed for motion control \cite{chen2012design}. However, actuators following this principle consume significant space within the spherical shell. The mass-imbalance principle involves unbalancing the centre of mass, as seen in the Glory by Javadi and Mojabi robot using axial rails attached to masses \cite{javadi2004introducing}. Another approach uses pendulum attachments by Liu et al. to shift the centre of mass and use internal space, with controllers exploring this principle \cite{liu2008stabilization}. Conservation of angular momentum is the basis for a third actuation principle, employing rotating internal gyroscopes or perpendicular rotor pairs \cite{bhattacharya2000spherical}. These designs, akin to a spinning ice skater, face challenges such as low torque production and sensitivity to external forces. Understanding these actuation types provides insights into the diverse mechanisms driving spherical robots. There have been attempts to create a hybrid driving mechanism using different principles of actuation \cite{schroll2008design,Tafrishi2019,arif2023design}. 

By assuming decoupled motion planning and dynamic control for trajectory tracking, the motion planning strategies can be applied similarly to S-CR, as outlined in Table \ref{Tab:expcase}. Spherical robots primarily involve a single contact point, and the degree of rotation depends on the actuator type. While the contact point can be determined through internal measurement units, precise haptic studies for accurately positioning contact points, especially when the robot navigates ramps or uneven terrains, remain an open problem in this research. However, potential solutions involve estimating contact points using camera/LiDAR information for navigation \cite{morad2018planning} through multiple agents. There is interest in exploring the underactuation \cite{he2019underactuated,wang2024rollbot} of the internal unit to maximize rotational capabilities by creating admissible paths through rolling contact kinematics, as developed for a rotational disk-actuated rolling robot in simulation studies \cite{morinaga2014motion}. This approach integrate the robot's dynamics with contact motion kinematics and there are challenges in inertial-coupling singularities \cite{spong1994partial,bergerman1995dynamic,tafrishi2020singularity} specially in rotating mass-point systems \cite{arai1991position,tafrishi2020singularity,tafrishi2021inverse} which could have limitations in robot mechanism design. Tafrishi et al. showed rolling robots has similar configuration singularities as the linked manipulators \cite{tafrishi2020singularity}.

The model of rolling robots using Montana kinematics through curvature models has been extensively explored \cite{svinin2008motion,li1990motion,marigo2000rolling,lynch2003control,tafrishi2021darboux,bai2018dynamic}. When it comes to arbitrary ground surfaces, there has been a lack of comprehensive studies focusing on curvature-based modeling rather than Cartesian space modeling for predetermined rough terrain \cite{moazami2019kinematics}. However, we can determine the rolling sphere $U_{f}=\{(u_{f},v_{f})|\;$$-\pi<u_{f}<\pi,$ $-\pi<v_{f}<\pi\}$ with radius $R_f$ on a generalized surface $U_o$ on the ground by considering the normal $(k^o_{nu},k^o_{nv})$, geodesic curvature $(k^o_{gu},k^o_{gv})$, and geodesic torsion $(\tau^o_{gu},\tau^o_{gv})$ of the surfaces to find (\ref{Eq:Montanakinemitcstransformed}) using
\begin{align}
 & \uvec{K}_{f}=\left[\begin{array}{cc}
  	\frac{1}{R_f} & 0\\
  	0 & \frac{1}{R_f}
  \end{array}\right], \uvec{K}_{o}=\left[\begin{array}{cc}
  	k^{o}_{nu} (u_o,v_o)& \tau^{o}_{gu}(u_o,v_o)\\
  	\tau^{o}_{gv} (u_o,v_o)& 	k^{o}_{nv}(u_o,v_o)
  \end{array}\right]  \nonumber\\
& \uvec{T}_{f}= \left[\begin{array}{c}
  	0 \\
  	-\tan v_{f} / R_f
  \end{array}\right] , \uvec{T}_{o} = \left[\begin{array}{c}
  	k^{o}_{gu} (u_o,v_o) \\
  	k^{o}_{gv} (u_o,v_o)
  \end{array}\right],
\label{Eq:curvaturepropertiesofrollingrobotM}
\end{align}
The curvature properties expressed in \eqref{Eq:curvaturepropertiesofrollingrobotM} show how the off-rotating surface $U_{f,i}$ can serve as the rolling robot, with the main object acting as the ground surface $U_{o}$. This formulation is adaptable for any arbitrary surface, given the contacted curvature, with symmetric surfaces like a planar ground having zero curvature. However, a significant challenge arises when dealing with variable curvature (or changing contact radius $R_{o}(u_o,v_o)$) through generalized local coordinates. Especially when a rolling sphere traverses continuously changing surfaces, the curvature properties need to be time-dependent or angular state-dependent $(u_o,v_o)$. This dependency complicates the kinematic model as it introduces changes throughout due to its reliance on the derivatives of the main curvature parameters which needs further studies, especially in the area of accessibility for generalizable path planning.

Ballbots typically utilize a standard fixed actuation type, employing multiple motors and omni-wheels with tangential surfaces in contact with the sphere \cite{morad2018planning,xuan2023intelligent}. While the rotating ball primarily functions as an actuator between the platform and the ground, its rotational dynamics are often disregarded in motion planning studies. The common objective is to maintain platform balance (or respect to the centre of the ball) and move it in translational planar axes to a desired location. The key idea had an omnidirectional mobile platform that transfers multiple wheel actuators' rotational momentum to crease arbitrary rotation of a rolling sandwiched ball. This type of platform also has found application in assistive robot systems like unidirectional wheelchairs standing on top of the ball \cite{song2024driving}.

For ballot systems with a specific configuration, such as those with $n$ omni-wheels actuators and a ground interface of $n+1$, considering motion planning from the sphere's perspective becomes crucial which was not considered before in any study. The contact motion kinematics for this problem can be found from (\ref{Eq:FinalGeneralKinematics}) by defining the normal, geodesic curvatures and geodesic torsion of flywheels surface $U_{fi}$ by $\left\{ \uvec{T}_{f1},\uvec{K}_{f1}\cdots\uvec{T}_{fi},\uvec{K}_{fi}\right\}$, the intermediate ball surface $U_{o}$ by $\left\{ \uvec{T}_{oi},\uvec{K}_{oi}\right\}$ and flat ground surface $U_{f0}$ by $\left\{ \uvec{T}_{f0},\uvec{K}_{f0}\right\}$ as follows
\begin{align} 
& \uvec{K}_{oi}=\left[\begin{array}{cc}
  	\frac{1}{R_o} & 0\\
  	0 & \frac{1}{R_o}
  \end{array}\right],\;\uvec{T}_{oi} =\left[\begin{array}{c}
  	0 \\
  	-\tan v_o / R_o
  \end{array}\right],  \nonumber \\
  & \uvec{K}_{fi}=\left[\begin{array}{cc}
  	\frac{1}{R_{fi}} & 0\\
  	0 & \frac{1}{R_{fi}}
  \end{array}\right],\; \uvec{T}_{fi}= \left[\begin{array}{c}
  	0 \\
  	-\tan \zeta  / R_{fi}
  \end{array}\right]\nonumber\\
 &\uvec{K}_{f0}=\left[\begin{array}{cc}
  	0 & 0\\
  	0 &  0
  \end{array}\right], \uvec{T}_{f0}=  \left[\begin{array}{c}
  	0 \\
  	0
  \end{array}\right] .
\end{align}
Note that the curvature properties $\left\{ \uvec{T}_{oi},\uvec{K}_{oi}\right\}$ of rotating disks $U_{f,i}$ exhibit singular configurations, preventing the direct application of Montana kinematics. Therefore, we have shown singular configurations and proposed a method to avoid singularities from the geometrical form, as detailed in Appendix \ref{Appex:AvoidSingularMontana}. The ground properties can be changed to a similar format of $\left\{\uvec{K}_{f0}(u_{f0},v_{f0}),\uvec{T}_{f0}(u_{f0},v_{f0})\right\}$ in (\ref{Eq:curvaturepropertiesofrollingrobotM}) if an arbitrary surface model is considered.

To date, a noticeable gap exists in the literature regarding this unique motion planning problem. One might question why considering the contact point of the sphere and the ground is crucial in motion planning. However, this becomes significantly important when seeking an optimal solution to determine trajectories on uneven trains. The sphere beneath the platform serves as the main rotating surface, with the motors primarily responsible for efficient balancing and displacements. Trajectories generated on the sphere illustrate how the robot traverses, introducing complications due to constraints from the contacts of motor wheels, necessitating their specific locations on the upper hemisphere \cite{lauwers2006dynamically,cai2019kinematic}. Utilizing S-CR methods, such as ball-plate methods, is a viable approach, especially when the friction between the sphere and the ground surpasses that between the actuator and the sphere. For example, while the method by Date et al. \cite{date2004simultaneous} can be applied to construct a suitable trajectory toward desired goals (upper plane as the cart with disk actuators, sandwiched sphere as ball of ballot, and ground as planar surface), it's crucial to note the requirement for a trajectory tracking controller to maintain the motor contacts in the upper hemisphere. This ensures that the platform remains upright, presenting an intriguing challenge that warrants in-depth study. Most similar studies \cite{morad2018planning,cai2019kinematic, xuan2023intelligent} focus on the two angular states rather than the local coordinates of each motor contact point ($\uvec{u}_{o,i},\uvec{u}_{f,i}$). This new perspective better explains the situation and manoeuvring area, treating each motor's wheel contact point as two state variables and transforming the problem from two two-variable states to $4(n+1)$ states.
 
\subsection{Dexterous Manipulation}
Rolling contacts are pivotal in a variety of scenarios, particularly within grasping mechanisms, where they play a significant role in enabling dexterous manipulation. As outlined in Section \ref{Sec:PathPlanning}, rolling contact kinematics can be understood in two primary cases: fixed fingertips, which require adjustments in finger joint rotation and movement to modify the contact point on the grasped object, and rotating fingertips, where the contact point undergoes infinite rotation without necessitating changes in finger joint actuation \cite{tahara2012externally,yuan2020designIROS}. Table \ref{Tab:expcase} summarizes the characteristics of these rolling contact forms. Additionally, Ozawa and Tahara provided an insightful review on rolling grasping and the utilization of control theory to maintain finger stability on objects \cite{ozawa2017grasp}. In cases where the fingertips do not have infinite rotation and are fixed within a limited area of touch, the complexity of the problem regarding the contact point is challenging, as it requires high friction at the contact point while the object orientation may not be highly significant. This study focuses on scenarios where the contact point is established, and $(n-1)$ number of fingertips are expected to change their configuration while maintaining continuous contact. From a modeling perspective (\ref{Eq:FinalGeneralKinematics}), a similar hemisphere $U_{f,i}=\{(u_{f,i},v_{f,i})|\;$$-\pi/2<u_{f,i}<\pi/2,$ $-\pi<v_{f,i}<\pi\}$ is typically assumed as $i$-th fingertip surface, while the rotating object is represented by a generalized convex surface model
\begin{align}
 & \uvec{K}_{fi}=\left[\begin{array}{cc}
  	\frac{1}{R_{fi}} & 0\\
  	0 & \frac{1}{R_{fi}}
  \end{array}\right],\; \uvec{T}_{fi}= \left[\begin{array}{c}
  	0 \\
  	-\tan v_{fi} / R_{fi}
  \end{array}\right],    \nonumber\\
& \uvec{T}_{oi} = \left[\begin{array}{c}
  	k^{o}_{gu,i} (u_o,v_o) \\
  	k^{o}_{gv,i} (u_o,v_o)
  \end{array}\right], \; \uvec{K}_{oi}=\left[\begin{array}{cc}
  	k^{o}_{nu,i} (u_o,v_o)& \tau^{o}_{gu,i}(u_o,v_o)\\
  	\tau^{o}_{gv,i} (u_o,v_o)& 	k^{o}_{nv,i}(u_o,v_o)
  \end{array}\right].
\label{Eq:curvaturepropertiesofrollingrobot}
\end{align}
These properties limit the planning strategies developed in S-CR to only certain planners that the constraining the area of contact is allowed e.g., Li and Canny \cite{li1990motion}, Svinin and Hose \cite{svinin2008motion} and Tafrishi et al. \cite{tafrishi2023geometric}.

\begin{figure}[t!]
	\centering
	\includegraphics[scale=.5]{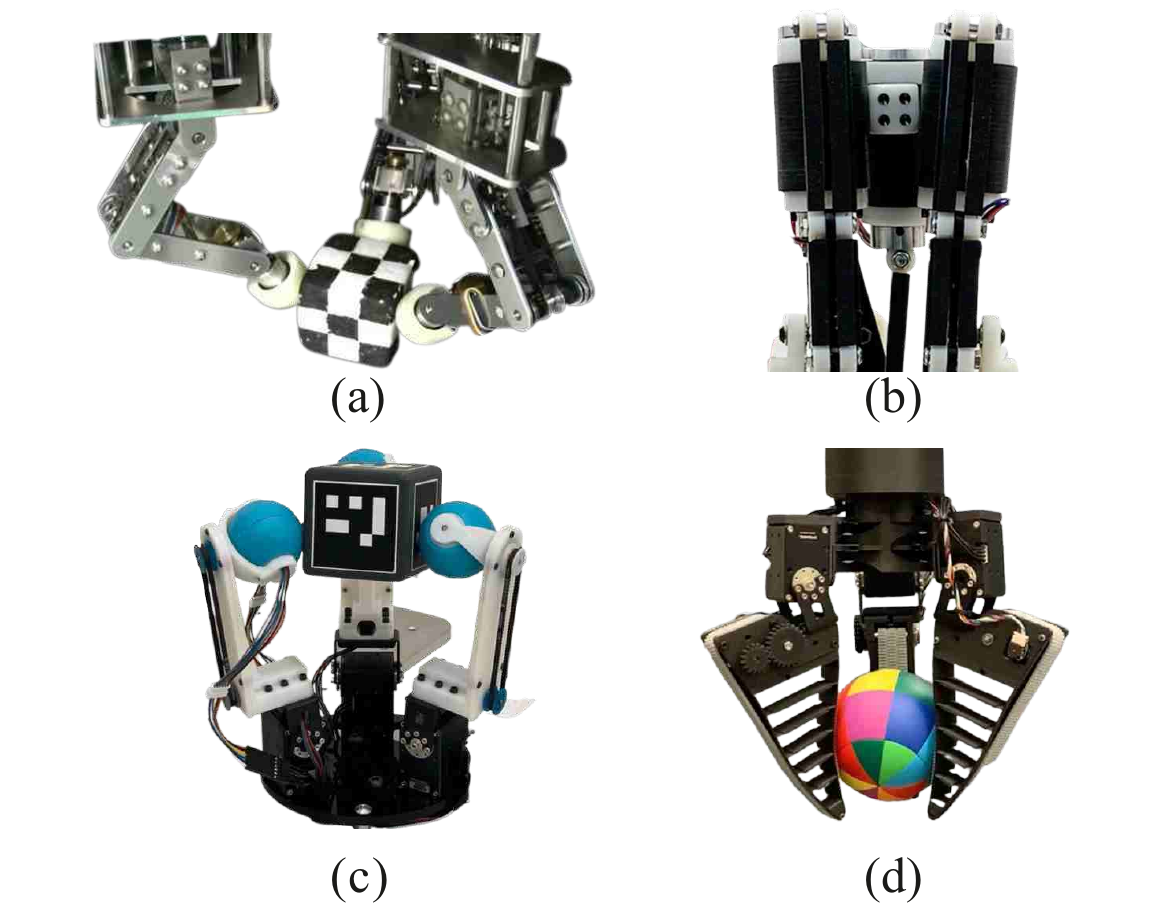}
	\caption{Designed agile grasping mechanisms with rotating fingertips for dexterous manipulation. a) 1-DoF Fingertip \cite{tahara2012externally} b) The rolling bar fingertip \cite{yuan2020designICRA} c) Rolling spherical fingertip \cite{yuan2020designIROS} d) Compliant rail fingertip \cite{cai2023hand}.}
\label{Fig:Grasping_Mechanism}
\end{figure}

There have been recent advancements in the development of mechanisms with infinitely rotating fingertips, as illustrated in Figure \ref{Fig:Grasping_Mechanism}. The novelty lies in leveraging the continuous rotational motion capability of these mechanisms, enabling more dexterous manipulation of objects without the need to release the grasp. The unconventional manipulation ability provided by mechanisms with infinitely rotating fingertips, absent in biological grasping hands like those of humans, allows robots to rotate objects dexterously without the risk of instability associated with releasing the grasp. These advancements hold great promise for agile grasping mechanisms, facilitating faster and more efficient object manipulation in various industrial applications. The development of agile fingertips has seen notable progress, starting with Tahara et al. presentation of a 1-DoF mechanism \cite{tahara2012externally}.  Recent alternative designs by Yuan et al. include the incorporation of a rolling bar \cite{yuan2020designICRA} or a spherical fingertip \cite{yuan2020designIROS}, upgrading the DoF to two. Another approach proposed by Cai et al. involves increasing grasping contact with compliant surfaces, albeit limiting the fingertip's DoF to one \cite{cai2023hand}. However, the challenge of increasing the degree of rotation while maintaining compliance and flexibility to handle different objects remains an open problem in the field of grasping agile fingertips. 

Understanding the precise point or area of contact remains a significant challenge in robotic haptics, with extensive research focused on developing sensing technologies for detecting the form/geometry and location of contact on grasped objects \cite{navarro2021proximity}. This issue is crucial for motion planning in dexterous manipulation tasks involving object handling, necessitating a deeper understanding of object or compliant fingertips \cite{ghafoor2004stiffness} properties such as stiffness and surface characteristics (e.g., softness). It necessitates incorporating these models into (\ref{Eq:Montanakinemitcstransformed}) and (\ref{Eq:FinalGeneralKinematics}).   
Current research efforts focus on integrating advanced sensor technologies, such as visual-tactile designs \cite{yuan2023tactile}, with enhanced agility in fingertip areas, alongside contact-impact impedance control models \cite{AritaHirata2023}. These advancements aim to enable more precise and robust dexterous manipulation tasks.



\subsection{Nonprehensile Manipulation} 
\label{nonprehensilemaniapplicaiton}
The role of contact kinematics in nonprehensile manipulation scenarios presents an interesting yet challenging task, both in motion planning and technology-wise. These tasks require rigorous modelling and high-frequency sensors/actuators to create robust handling since the motions are sparse and quick, such as balancing or throwing actions that fall under nonprehensile problems. Ruggiero et al. \cite{ruggiero2018nonprehensile} provided insightful discussions on how contact potential changes depending on task requirements and object surface properties.

The coupling between robot and part subsystems occurs through contact and impact laws. Controlling nonprehensile manipulation systems heavily relies on friction and impact laws \cite{lynch2003control}, with much of the complexity stemming from these factors. The contact law is inherently non-smooth due to changes in robot-part (or hand-object) contact geometry and variations in friction law, transitioning between slipping and sticking contact. Each contact mode, characterized by the set of contacts and their current state (slipping or sticking), entails its own set of dynamic equations. Transitions in contact mode result in changes in these equations, some of which are controllable while others are not. Consequently, nonprehensile systems encounter challenges akin to those in general hybrid control systems.

Balancing, a continuous contact mode action reminiscent of rolling a ball in the human palm, has inspired numerous studies on ball and beam problems \cite{ruggiero2018nonprehensile,woordrufLynch2023TRO} and disk on disk \cite{DiskonDiskLynch2013}. The problem is often regarded more as a control problem than a motion planning one due to the prevalence of dynamic issues such as sliding and gravity which makes it harder to find admissible paths online form. Notably, there has been limited research on rolling soft objects that deform while rotating, such as dough, which necessitates carefully planned trajectories for effective manipulation. Juggling, another contact mode, involves shorter time-lapses where the palm maintains contact with the ball while rolling and moving it through the air \cite{lynch2003control}. Conversely, actions like batting and catching rely more on impact laws \cite{batz2010dynamic}, thus minimizing the complexity of contact planning. However, similar to the motion of legged robots, planning for the contact of rotating objects or hitting a ball with a baseball bat requires robust and real-time planners capable of addressing momentum, aerodynamics and impact predition equations \cite{ruggiero2018nonprehensile}. There are more complex real-world tasks, such as flipping a pancake, tool manipulation, or foot placement in legged robots, which fall under nonprehensile manipulation as open problems \cite{chuah2019bi}. In legged robots, foot placement requires continuous periodic engagement with the ground to maintain dynamic stability \cite{chuah2019bi,Hutter2020}. However, effective contact planning can enhance the robot's agility and ability to navigate challenging terrains. Aforementioned object manipulation tasks, on the other hand, are often addressed without considering the motion planning problem, using learning methods trained on discrete hybrid human-centric manipulations \cite{yu2022dexterous,pfanne2022hand}.
    \begin{figure}[t!]
	\centering
	\includegraphics[scale=.5]{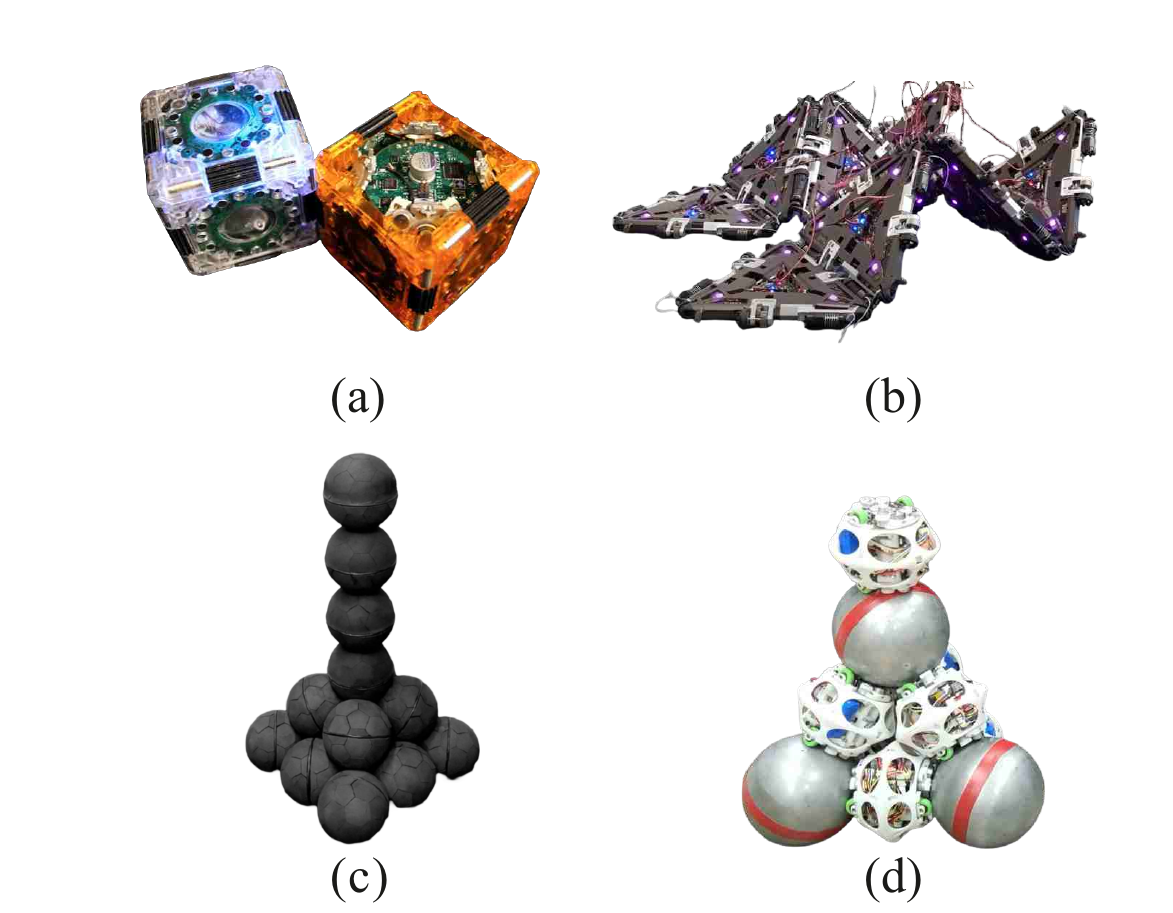}
	\caption{Developed reconfigurable robots with rotational contact points. a) 3D M-Blocks modular robot \cite{romanishin20153d} b) Mori morphological modular polygon robot \cite{belke2017mori,belke2023morphological} c) Spherical/Disk rotational modules \cite{zhong2022kin,Wiltshire2024disknovel} d) The intermediate modules with passive balls \cite{tu2023configuration}.}
\label{Fig:Reconfigurable_Mechanism}
\end{figure}



\subsection{Reconfigurable Swarm Robotics}

The most complex yet interesting group of robots is the category known as \textit{Reconfigurable Robots}, illustrated in Figure \ref{Fig:Reconfigurable_Mechanism}. These robots are envisioned as versatile swarms of modular platforms capable of changing shape and functionality to perform various tasks \cite{moubarak2012modular}. The concept involves creating robots that can execute different types of locomotion, such as walking, rolling, or spiralling motions, as well as manipulating their environment \cite{seo2019modular}. Extensive research in this field has focused on developing modular systems, particularly in swarm robotics. The concept involves the development of either dependent interconnected (Figure \ref{Fig:Reconfigurable_Mechanism} (a)-(b)) or independent attachable (Figure \ref{Fig:Reconfigurable_Mechanism} (c)-(d)) modular platforms capable of robust operation in diverse environments. There are still challenges in effectively coupling these modules, which imposes significant limitations on the overall capabilities of the robot \cite{romanishin20153d,belke2017mori,Wiltshire2024disknovel}.

The significance of the contact point was not as pronounced in earlier designs, as depicted in Figure \ref{Fig:Reconfigurable_Mechanism} (a)-(b), where it primarily governed a single angular rotation and remained mostly locked without traversing on top/surface of other modules \cite{romanishin20153d,belke2017mori,belke2023morphological}. However, recent advancements in extending the Degrees of Freedom (DoF) in each module's rotation, as shown in Figure \ref{Fig:Reconfigurable_Mechanism} (c)-(d), have introduced a new challenge of rolling contact kinematics, where each spherical module potentially interacts with more than one module through rotation atop each other \cite{zhong2022kin}, complicating the execution of motion planning strategies. Nonetheless, the internal driving unit is yet to be optimized to enable continuous rotation without sacrificing connection integrity or the ability to exert high torque on each coupling point to support and transport the interconnected modules. Attempts to address this challenge include externalizing the main actuators outside the spherical body; however, this approach tends to negate the advantages of a closed spherical body with internalized actuators \cite{tu2023configuration}.

In these robot platforms, the M-CR motion planning should be discussed while considering the geometry of the main body, which may be fixed by the robot's design, along with other existing contacts that are in similar form. However, there exists a coupling relation when planning each group of rotating convex shapes, such as spheres, which might affect others that are directly/indirectly dependent. The contact problem can be looked for $n$ convex surface modules that could have potential contact numbers ranging $\left[n-1,2(n-1)\right]$ for $n>2$. The path planning can be approached in two aspects: \textit{hierarchical form} and \textit{sub-group form}. In the former hierarchical form, certain groups of particles (convex surfaces), simplified as spheres, are prioritized over others, and planning begins with these high-priority spheres, with iteration proceeding downward based on ordered computation. For example, Luo et al. presented a tree-based branch-and-bound strategy for contact planning involving various passive shapes \cite{luo2022auto}. The later solution as the sub-group form involves either randomly or specifically choosing the most important group, depending on changing conditions about those particles, and then spreading around other neighbouring sub-groups with certain cost functions or convergence rules. Some recent shape formation researches with prioritised and decentralized planning strategies were studied in swarm robotics \cite{abujabal2023comprehensive}. In one practical example where particles act as manipulators (some particles function as links, others as joints, and some as the end-effector), our priority can be looked as the sub-group of particles at the end-effector due to their interaction with the environment. This becomes even more critical in complex scenarios involving direct human-robot interaction. Conversely, the process goes to other dependent connected particles as remaining sub-groups until it reaches the particles at the base. This sub-group strategy might not be easy to apply for the hierarchical form. Moreover, the issue becomes more complicated when there is more than one point of interaction, such as the end-effector, as there will be potential points where certain end-effector points might change simultaneously, requiring the dependent other changing sub-group to adapt to patterns with more than one change. An example illustrating this complexity can be observed in an octopus with multiple tentacles. The concept is simplified if each tentacle is dependent on the base, but when multiple tentacles are connected to each other, each branch may have a continuous effect on other procedures. This means the branch search methods are combined with M-CR kinematics which makes it more intricate without even considering the dynamics. 

The field of reconfigurable rolling robots also introduces another challenge: effectively integrating M-CR kinematics into \textit{shape planning}. This involves determining the traversal path for each particle in a reconfigurable robot to achieve a specific shape. It's important to note that this challenge can be studied independently of the previously mentioned problem of planning order (hierarchical or sub-group). This assumption arises from viewing the problem as a statement of passive object fitting, where the states of rolling particles can be disregarded, reducing the problem to fitting the particles into the desired shape. However, the order of motion planning becomes relevant when there is a need for active control or shape change. This aspect is crucial for making reconfigurable robots resemble existing active platforms or robots, such as cranes, wheelchairs, manipulators, humanoids, bio-insipred systems etc.

The contact kinematics of M-CR in Equation (\ref{Eq:Montanakinemitcstransformed}) for reconfigurable robots can be simplified to a problem similar to agile fingertip grasping. In this scenario, the central object serves as the main iterated object, while the other objects act as rotating off-rolling modules (analogous to fingertips in grasping). Due to the uniform similarity of all modules, and if considered as spheres, the same curvatures ($(\uvec{K}_{oi}, \uvec{T}_{oi})$ are spherical geometries in this problem) in Equation (\ref{Eq:curvaturepropertiesofrollingrobot}) can be utilized to obtain multi-contact rolling (M-CR) kinematics. However, this computation should happen in $n$ iterations with a computation complexity of $n!$ for each grouped module kinematics and be solved based on discussed order forms (hierarchical or sub-group).

\subsection{Micro/Nano Particle Manipulation}
Particle manipulation at the micro or nano level requires specific design considerations for sliding and rolling particles \cite{zhang2009autonomous,pawashe2011two,ali2016fabrication,bozuyuk2022reduced,yang2019microwheels,law2023micro}. Most particles are typically manipulated and considered one-by-one, placing them in the realm of single-contact rolling (S-CR) in most cases. Therefore, S-CR path planning strategies can be employed to utilize admissible trajectories.

Micro/nano particle manipulation has a wide range of applications across various fields. In biomedical engineering, precise control of micro/nano particles is crucial for targeted drug delivery in bloodstreams, where particles can be directed to specific sites within the body to release medication with high precision \cite{ali2016fabrication}. This technology is also pivotal in developing advanced diagnostic tools, such as lab-on-a-chip devices, where micro/nano particles can be manipulated to interact with biological samples for rapid and accurate testing \cite{bozuyuk2022reduced}. In materials science, micro/nano particle manipulation enables the fabrication of novel materials with tailored properties. By precisely positioning particles, researchers can create materials with unique mechanical, electrical, or optical characteristics \cite{zhang2009autonomous}. This level of control is essential for developing next-generation sensors, actuators, and other micro-scale devices \cite{law2023micro}. In robotics, micro particle manipulation is crucial for developing micro-robots capable of performing tasks in confined or hazardous environments. These robots can navigate through small spaces, perform precise manipulations, and even assemble microstructures, making them valuable for applications ranging from industrial manufacturing to space exploration \cite{pawashe2011two,law2023micro}.

The kinematics of sliding and rolling particles are particularly important and warrant discussion on how they may be affected. For instance, the coupling of dynamic factors at the nano/micro level becomes necessary. Actuation methods often involve external forces such as physical needles to move the spherical particles \cite{zhang2009autonomous}, magnetic fields to rotate them \cite{pawashe2011two,ali2016fabrication}, or displacement through fluid flow \cite{bozuyuk2022reduced}. Consequently, kinematics should be coupled with the dynamics of actuation for proper manipulation, taking into account the constraints and forces specific to the micro/nano scale. Sliding becomes crucial in this context. However, detecting sliding poses challenges as sensory feedback is primarily visual, which may have inherent delays and errors depending on particle size. This introduces significant complexities in control strategies, necessitating high-frequency and high-precision sensors to mitigate these issues. Accurate environmental modeling can alleviate some of these challenges by predicting particle behavior and compensating for feedback delays.

In particle manipulation environments, external disturbances are often minimized, allowing for highly controlled conditions. This controlled setting facilitates the precise execution of complex tasks. For instance, understanding the interaction forces at play and the influence of surface properties is critical for effective particle manipulation.

\section{Conclusion}
In this survey paper, we have delved into the intricate domain of rolling contact motion planning across various robotic applications, highlighting both advancements and persistent challenges in this field. A significant contribution of this survey is the connection between existing problems and rolling contact kinematics, shedding light on novel approaches and methodologies for addressing complex robotic tasks. We proposed a hypothesis for solving complex multi-contact kinematics by leveraging developed path-planning methods for single-contact kinematics.

We also summarized the different characteristics of single and multi-contact kinematics in various robotic applications to highlight open problems. Starting with rolling robots and ballbots, we explored the existing challenges and planning approaches for pure rolling and spin-rolling motions, connecting them to the dynamics based on the environment or internal actuators. We then discussed dexterous manipulation regarding multi-contact motion planning and its relevance to conventional fixed and rolling agile fingertips. The importance of contact motion planning for nonprehensile manipulations was also emphasized, relating to challenges such as balancing objects, juggling, or catching a ball. Similar issues were noted in foot placement for legged robots, which is crucial in uneven terrains. We introduced new issues found in reconfigurable robots, where modular body displacement presents challenges similar to multi-finger manipulation, further extending the complexity to shape planning. Integrating multi-contact motion planning with advanced sensing technologies remains an ongoing challenge, essential for real-time adaptation and robust autonomous operation. Finally, we addressed material science applications, particularly micro/nano particle manipulation. Controlling spherical or convex particles is crucial for developing novel materials and surface forms, a hot topic in the field. In summary, while significant progress has been made, ongoing research is needed to address these multifaceted challenges, paving the way for future breakthroughs in this dynamic and rapidly evolving field.


\section*{DISCLOSURE STATEMENT}
The authors are not aware of any affiliations, memberships, funding, or financial holdings that
might be perceived as affecting the objectivity of this review. 

\section*{ACKNOWLEDGMENTS}
This work was supported by the Royal Society research grant under Grant \text{RGS\textbackslash R2\textbackslash 242234}.

%

\bibliographystyle{ar-style3}
\bibliography{References}

\newpage
\begin{appendices}
\section{Preliminaries for the Multi-contact Kinematics }
\label{MultiFingerComputAppendix}
This section derives the multi-contact kinematics presented in (\ref{Eq:FinalGeneralKinematics}) using generalized Montana kinematics (\ref{Eq:Montanakinemitcstransformed}). In this problem, it is assumed as shown in Figure \ref{Fig:PathPlanningMultifingergraspcdr}, there is a central main object $U_o$ in contact with other off-rotating objects $U_{f,i}$. At first, the relative rotation angular velocities on contact frames are already defined by 
\begin{align}
\left[\begin{array}{c}
\bm{\omega}_{rel,1} \nonumber \\
 \vdots \nonumber\\
\bm{\omega}_{rel,i} \nonumber\\
\uvec{v}_{rel,1} \nonumber\\
\vdots \nonumber \\
\uvec{v}_{rel,i} 
\end{array}\right] = \left[\begin{array}{c}
 \bm\omega_{f,1}-\bm\omega_{o} \nonumber \\
 \vdots \nonumber\\
 \bm\omega_{f,i}-\bm\omega_{o} \nonumber\\
\uvec{v}_{f,1} - \uvec{v}_{o}  \nonumber\\
\vdots \nonumber \\
\uvec{v}_{f,i} - \uvec{v}_{o} 
\end{array}\right].
\end{align}
An arbitrary off-rotating surface $U_{f,i}$ on the contact $\Sigma_{C,i}$ with the main rotating object $U_o$ has a reference frame $\Sigma_O$. The angular and linear velocity relations based on the Euler angles of the $i$-th object as $\bm{\Psi}_i=(\theta_{x,i},\theta_{y,i},\theta_{z,i})$ and the Cartesian position $\uvec{X}_i=(x_i,y_i,z_i)$, which extend Sarkar's work \cite{sarkar1997control,sarkar1996velocity}, are derived as:
\begin{eqnarray}
     \bm{\omega}_{f,i} &=&\bm{\omega}_{r,f_i}=\uvec{R}_{c_o,c_i}\uvec{R}_{c_i,f_i}\uvec{R}_{f_i,r} \tilde{\uvec{R}}_{f,i} \dot{\bm{\Psi}}_i=\uvec{H}_{f,i} \dot{\bm{\Psi}}_i, \nonumber \\
     \label{Eq:AngularRelationBasiangularvel}
    \bm{\omega}_{o} &= &\bm{\omega}_{r,o}=\uvec{R}_{c_o,o}\;\uvec{R}_{o,r} \tilde{\uvec{R}}_o\; \dot{\bm{\Psi}}_o =\uvec{H}_{o} \dot{\bm{\Psi}}_o, \\
    \uvec{v}_{f,i} & =& \uvec{v}_{r,p_{i}}=\uvec{R}_{c_o,c_i}\uvec{R}_{c_i,f_i}\uvec{R}_{f_i,r} \dot{\uvec{X}}_{f,i} +\uvec{H}_{f,i}\dot{\bm{\Psi}}_i \times \uvec{R}_{c_i,c_o}\uvec{R}_{c_i,f_i} \uvec{r}_{f_i,p_{i}} \nonumber \\
     &=& \uvec{R}'_{f,i} \dot{\uvec{X}}_{f,i} - \uvec{H}'_{f,i} \dot{\bm{\Psi}}_{f,i},  \nonumber  \\
    \uvec{v}_{o}& = & \uvec{v}_{r,p_o}= \uvec{R}_{c_o,o}\uvec{R}_{o,r}\dot{\uvec{X}}_{o}+ \uvec{H}_o \dot{\bm{\Psi}}_o \times \uvec{R}_{c_o,o} \uvec{r}_{o,p_o} = \uvec{R}'_{o} \dot{\uvec{X}}_{o} - \uvec{H}'_{o} \dot{\bm{\Psi}}_{o} ,
   \label{Eq:AngularRelationBasiclinearvel}
\end{eqnarray}
where collected rotational transformation matrices are 
\begin{align}
    & \uvec{H}_{f,i}  = \uvec{R}_{c_o,c_i}\;\uvec{R}_{c_i,f_i}\;\uvec{R}_{f_i,r}\; \tilde{\uvec{R}}_{f,i} , \uvec{H}_o = \uvec{R}_{c_o,o}\;\uvec{R}_{o,r}  \tilde{\uvec{R}}_o, \nonumber\\
    & \uvec{R}'_{f,i} = \uvec{R}_{c_o,c_i}\;\uvec{R}_{c_i,f_i}\;\uvec{R}_{f_i,r}, \uvec{R}'_o=\uvec{R}_{c_o,o}\;\uvec{R}_{o,r},  \nonumber\\ 
    & \uvec{H}'_{f,i} = \left[ \uvec{R}_{c_o,c_i}\;\uvec{R}_{c_i,f_i} \;\uvec{r}_{f_i,p_{i}} \times \right]  \uvec{H}_{f,i}, \uvec{H}'_o= \left[ \uvec{R}_{c_o,o} \uvec{r}_{o,p_o} \times \right]  \uvec{H}_ o.
    \label{Eq:Matrixcompactfoms}
\end{align}
The skew-symmetric radii for (\ref{Eq:Matrixcompactfoms}) the $i$-th off-rotating objects $[\uvec{r}_{f_i,p_{i}} \times ]$ and the main object $[\uvec{r}_{o,p_o}\times ]$ are expressed as
\begin{align}
& [\uvec{r}_{f_i,p_{i}} \times ]=\left[\begin{array}{ccc}
0 & -r_{z,f_i}& r_{y,f_i}\\
 r_{z,f_i} &0  &  -r_{x,f_i}\\
-r_{y,f_i} & r_{x,f_i} & 0
\end{array}\right],   [\uvec{r}_{o,p_o}\times ]= \left[\begin{array}{ccc}
0 & -r_{z,o}& r_{y,o}\\
 r_{z,o} &0  &  -r_{x,o}\\
-r_{y,o} & r_{x,o} & 0
\end{array}\right],   \uvec{R}_{c_o,c_i}= \left[\begin{array}{cc}
\uvec{R}_\psi & 0 \\
0  & -1
\end{array}\right].
\end{align}
It is important to note that the radii should be time- or coordinate-dependent if the surface curvature is changing, especially for soft or compliant object surfaces on either the main or off-rotating surfaces. Also, the transformations $\uvec{R}_{o,c_o}$ and $\uvec{R}_{f_i,c_i}$ depend on the utilized coordinate system for defining the local surface geometries. 

Then, by substituting the derived formulation with Euler transformation in (\ref{Eq:AngularRelationBasiangularvel})-(\ref{Eq:AngularRelationBasiclinearvel}) into (\ref{Eq:Angularrelativeveloeq}) and reordering them, we will derive our final form of the equations that give us the complete model of kinematics for the multi-contact (similarly shown in Equation (\ref{Eq:FinalGeneralKinematics})) as 
\begin{eqnarray}
\dot{\uvec{U}}=\left[\begin{array}{c}
\uvec{u}_{o,1}\\
\uvec{u}_{f,1}\\
\psi_1\\
\uvec{u}_{o,2}\\
\uvec{u}_{f,2}\\
\psi_2\\
\vdots \\
\uvec{u}_{o,i}\\
\uvec{u}_{f,i}\\
\psi_i\\
\end{array}\right]=\uvec{D}\dot{\uvec{q}}=\uvec{D}\left[\begin{array}{c}
\dot{\uvec{X}}_o\\
\dot{\bm{\Psi}}_o\\
\dot{\uvec{X}}_{f,1}\\
\dot{\bm{\Psi}}_{f,1}\\
\dot{\uvec{X}}_{f,2}\\
\dot{\bm{\Psi}}_{f,2}\\
\vdots \\
\dot{\uvec{X}}_{f,i}\\
\dot{\bm{\Psi}}_{f,i}\\
\end{array}\right],
\end{eqnarray}
where
\begin{align}
\uvec{D}=\left[\begin{array}{ccccc}
 & \uvec{D}_{f,1} & \uvec{0}  & \uvec{0} & \uvec{0}\\
  & \uvec{0} & \uvec{D}_{f,2} &\uvec{0} &\uvec{0}\\
\uvec{D}_o &   &   & \ddots  & \\
& \uvec{0} & \uvec{0} & \uvec{0} & \uvec{D}_{f,i}
\end{array}\right],
\end{align} 
while we have 
\begin{equation*}
\uvec{D}_o=\left[\begin{array}{cccccccc}
 ^1 \uvec{D}_{o,1}&^3 \uvec{D}_{o,1} & ^5 \uvec{D}_{o,1}& \cdots &  ^1 \uvec{D}_{o,i}&^3 \uvec{D}_{o,i} & ^5 \uvec{D}_{o,i} \\
^2 \uvec{D}_{o,1}&^4 \uvec{D}_{o,1}  & ^6 \uvec{D}_{o,1} & \cdots &  ^2 \uvec{D}_{o,i}&^4 \uvec{D}_{o,i} & ^6 \uvec{D}_{o,i}
\end{array}\right]^T
\end{equation*} 
\begin{align}
& \uvec{D}_{f,1}= \left[\begin{array}{cc}
^1 \uvec{D}_{f,1} & ^2 \uvec{D}_{f,1} \\
^3 \uvec{D}_{f,1}  & ^4 \uvec{D}_{f,1} \\
^5 \uvec{D}_{f,1} & ^6 \uvec{D}_{f,1}
\end{array}\right] \cdots \uvec{D}_{f,i}= \left[\begin{array}{cc}
^1 \uvec{D}_{f,i} & ^2 \uvec{D}_{f,i} \\
^3 \uvec{D}_{f,i}  & ^4 \uvec{D}_{f,i} \\
^5 \uvec{D}_{f,i} & ^6 \uvec{D}_{f,i}
\end{array}\right].
\end{align}
These matrices are defined by
\begin{align}
 & ^1 \uvec{D}_{o,1}=-\bm{\Upsilon}_{o1}\tilde{\uvec{K}}_{f1}\uvec{E}_3\uvec{R}'_{o1},\cdots,^1 \uvec{D}_{o,i}= -\bm{\Upsilon}_{oi}\tilde{\uvec{K}}_{fi}\uvec{E}_3\uvec{R}'_{oi}, \nonumber\\
& ^2 \uvec{D}_{o,1}=\bm{\Upsilon}_{o1} \left (\uvec{E}_2 \uvec{H}_{o,1} + \tilde{\uvec{K}}_{f1}\uvec{E}_3\uvec{H}'_{o1} \right),\cdots, \nonumber\\
 & ^2 \uvec{D}_{o,i}= \bm{\Upsilon}_{oi} \left (\uvec{E}_2 \uvec{H}_{o,i} + \tilde{\uvec{K}}_{fi}\uvec{E}_3\uvec{H}'_{oi} \right) \nonumber\\
 & ^3 \uvec{D}_{o,1}= \bm{\Upsilon}_{f1}\uvec{K}_{o1}\uvec{E}_3 \uvec{R}'_{o1}, \cdots,    ^3 \uvec{D}_{o,i}= \bm{\Upsilon}_{fi}\uvec{K}_{oi}\uvec{E}_3 \uvec{R}'_{oi}, \nonumber \\
  & ^4 \uvec{D}_{o,1}=\bm{\Upsilon}_{f1} \left (\uvec{E}_2 \uvec{H}_{o,1} - \uvec{K}_{o1}\uvec{E}_3\uvec{H}'_{o1} \right),\cdots, \nonumber\\
 & ^4 \uvec{D}_{o,i}= \bm{\Upsilon}_{fi} \left (\uvec{E}_2 \uvec{H}_{o,i} - \uvec{K}_{oi}\uvec{E}_3\uvec{H}'_{oi} \right) \nonumber\\
 & ^5\uvec{D}_{o,1}=\uvec{T}_{f1} \uvec{M}_{f1} {^3}\uvec{D}_{o,1} + \uvec{T}_{o1} \uvec{M}_{o1} {^1}\uvec{D}_{o,1} \nonumber\\
  & ^5\uvec{D}_{o,i}=\uvec{T}_{fi} \uvec{M}_{fi} {^3}\uvec{D}_{o,i} + \uvec{T}_{oi} \uvec{M}_{oi} {^1}\uvec{D}_{o,i} \nonumber\\
 & ^6\uvec{D}_{o,1}=\uvec{T}_{f1} \uvec{M}_{f1} {^4}\uvec{D}_{o,1} + \uvec{T}_{o1} \uvec{M}_{o1} {^2}\uvec{D}_{o,1}-\uvec{E}_2\uvec{H}_{o1}\nonumber\\
  & ^6\uvec{D}_{o,i}=\uvec{T}_{fi} \uvec{M}_{fi} {^4}\uvec{D}_{o,i} + \uvec{T}_{oi} \uvec{M}_{oi} {^2}\uvec{D}_{o,i} -\uvec{E}_2\uvec{H}_{oi}
  \label{Eq:Domatricesdetails}
\end{align}
\begin{align}
 & ^1 \uvec{D}_{f,1}=\bm{\Upsilon}_{o1}\tilde{\uvec{K}}_{f1}\uvec{E}_3\uvec{R}'_{f1},\cdots, ^1 \uvec{D}_{o,i}=\bm{\Upsilon}_{oi}\tilde{\uvec{K}}_{fi}\uvec{E}_3\uvec{R}'_{fi}, \nonumber\\
 & ^2 \uvec{D}_{f,1}=-\bm{\Upsilon}_{o1} \left (\uvec{E}_2 \uvec{H}_{f,1} + \tilde{\uvec{K}}_{f1}\uvec{E}_3\uvec{H}'_{f1} \right),\cdots, \nonumber\\
 & ^2 \uvec{D}_{f,i}= -\bm{\Upsilon}_{oi} \left (\uvec{E}_2 \uvec{H}_{f,i} + \tilde{\uvec{K}}_{fi}\uvec{E}_3\uvec{H}'_{fi} \right) \nonumber\\
 & ^3 \uvec{D}_{f,1}= -\bm{\Upsilon}_{f1}\uvec{K}_{o1}\uvec{E}_3 \uvec{R}'_{f1}, \cdots,    ^3 \uvec{D}_{f,i}= -\bm{\Upsilon}_{fi}\uvec{K}_{oi}\uvec{E}_3 \uvec{R}'_{fi}, \nonumber \\
  & ^4 \uvec{D}_{f,1}=-\bm{\Upsilon}_{f1} \left (\uvec{E}_2 \uvec{H}_{f,1} - \uvec{K}_{o1}\uvec{E}_3\uvec{H}'_{f1} \right),\cdots, \nonumber\\
 & ^4 \uvec{D}_{f,i}= -\bm{\Upsilon}_{fi} \left (\uvec{E}_2 \uvec{H}_{f,i} - \uvec{K}_{oi}\uvec{E}_3\uvec{H}'_{fi} \right) \nonumber\\
 & ^5\uvec{D}_{f,1}=\uvec{T}_{f1} \uvec{M}_{f1} {^3}\uvec{D}_{f,1} + \uvec{T}_{o1} \uvec{M}_{o1} {^1}\uvec{D}_{f,1} \nonumber\\
  & ^5\uvec{D}_{f,i}=\uvec{T}_{fi} \uvec{M}_{fi} {^3}\uvec{D}_{f,i} + \uvec{T}_{oi} \uvec{M}_{oi} {^1}\uvec{D}_{f,i} \nonumber\\
 & ^6\uvec{D}_{f,1}=\uvec{T}_{f1} \uvec{M}_{f1} {^4}\uvec{D}_{f,1} + \uvec{T}_{o1} \uvec{M}_{o1} {^2}\uvec{D}_{f,1}+\uvec{E}_2\uvec{H}_{f1}\nonumber\\
  & ^6\uvec{D}_{f,i}=\uvec{T}_{fi} \uvec{M}_{fi} {^4}\uvec{D}_{f,i} + \uvec{T}_{oi} \uvec{M}_{oi} {^2}\uvec{D}_{f,i} +\uvec{E}_2\uvec{H}_{fi}
  \label{Eq:Dfmatricesdetails} 
\end{align}
where
\begin{align}
& \bm{\Upsilon}_{o1} = \uvec{M}^{-1}_{o1}   (\uvec{K}_{o1}+\tilde{\uvec{K}}_{f1})^{-1}, \nonumber\\ &\bm{\Upsilon}_{f1} = \uvec{M}^{-1}_{f1} \uvec{R}_{\psi,f1} (\uvec{K}_{o1}+\tilde{\uvec{K}}_{f1})^{-1}\nonumber \\
&\bm{\Upsilon}_{oi}= \uvec{M}^{-1}_{oi}  (\uvec{K}_{oi}+\tilde{\uvec{K}}_{fi})^{-1},\nonumber \\
&\bm{\Upsilon}_{fi}= \uvec{M}^{-1}_{fi} 
\uvec{R}_{\psi,fi} (\uvec{K}_{oi}+\tilde{\uvec{K}}_{fi})^{-1}.
\end{align}
\section{Avoiding Rolling Disk Singularities in Montana Kinematics}
\label{Appex:AvoidSingularMontana}
Singularities in Montana kinematics \cite{montana1988kinematics} can arise when the local coordinates with their assigned angular parameters are less than two (not a 2D manifold). This issue is particularly noticeable in scenarios like wheels on a surface, such as mobile robots, or multiple-wheel placements on a ball for ball-bot robots.

We propose a mathematical solution for avoiding singularities while obtaining the complete kinematic model to address this. We transform a uniform spherical surface to a disk, assuming $v_o=0$. In this case, the only angular rotation for a rotating disk (unicycle) is given by $\{\omega_x,\omega_z\}$, where $\omega_z$ represents the velocity of angular spinning/steering by $\psi$. This transformation results in the following coordinates:
 \begin{align}
		& f_{o} : U_o \rightarrow \mathbb{R}^3
		: c(u_o,0)\mapsto \big[-R_{o}\sin{u_{o}},0,-R_{o}\cos{u_{o}}\big],
	\label{Eq:Coordinategeneralsurface}
\end{align}
By performing the necessary operations to find the curvature values, we obtain the parameters:
\begin{align}
 &\uvec{K}_{o}=\left[\begin{array}{cc}
  	k^{o}_{nu} & \tau^{o}_{nu}\\
  	\tau^{o}_{nv} & 	k^{o}_{nv}
  \end{array}\right]=\left[\begin{array}{cc}
  	0 & \infty\\
  	0 & 0
  \end{array}\right],\uvec{T}_{o} =\left[\begin{array}{c}
  	k^{o}_{gu} \nonumber\\
  	k^{o}_{gv}
  \end{array}\right]= \left[\begin{array}{c}
  	0 \\
  	0
  \end{array}\right]. 
\end{align}
As it is clear, the curvature properties are not any more realizable, and it stops us from obtaining the Montana kinematics. To deal with it, we propose that there is $v_o\approx \zeta$ where $\zeta$ is infinitesimal angular orientation.  By doing this, the formulation allows us to have curvature parameters as 
\begin{align}
 &\uvec{K}_{o}=\left[\begin{array}{cc}
  	k^{o}_{nu} & \tau^{o}_{nu}\\
  	\tau^{o}_{nv} & 	k^{o}_{nv}
  \end{array}\right]=\left[\begin{array}{cc}
  	1/R_o & 0\\
  	0 & 1/R_o
  \end{array}\right],\uvec{T}_{o} =\left[\begin{array}{c}
  	k^{o}_{gu} \nonumber\\
  	k^{o}_{gv}
  \end{array}\right]= \left[\begin{array}{c}
  	\tan(\zeta)/R_o \\
  	0
  \end{array}\right].
\end{align}
Subsequently, the kinematics of Montana are transformed as follows:
  \begin{eqnarray}
   \left[\begin{array}{c}
\dot{u}_o\\
\dot{v}_o \\
\dot{u}_f \\
\dot{\psi}
\end{array}\right] &= &  \left[\begin{array}{cc}
R_o &0\\
- R_o \tan\psi  &0\\
-1/\sin\psi &0\\
0 & 1
\end{array}\right] \left[\begin{array}{c}
 \omega_y\\
\omega_z
\end{array}\right]
\label{Eq:SimplifiedDiskonPlaneMon}
 \end{eqnarray}
This is a disk-on-plane scenario to clearly show the effect of the disk without any curvature properties of the second contact surface (plane).

This representation, based on our proposed assumption, allows for the incorporation of multiple disks into the complete kinetics. This is particularly important in the design of mobile robots, ball-bots, disk-based grasping mechanisms, and more.
\end{appendices}

\end{document}